\documentclass{article}

\usepackage[preprint]{neurips_2026}
\usepackage[utf8]{inputenc} 
\usepackage[T1]{fontenc}    
\usepackage{hyperref}       
\usepackage{url}            
\usepackage{booktabs}       
\usepackage{amsfonts}       
\usepackage{nicefrac}       
\usepackage{microtype}      
\usepackage{xcolor}         
\usepackage{amsmath}
\usepackage{graphicx}
\usepackage{amsthm}
\usepackage{amssymb}
\usepackage{amsmath}
\usepackage[capitalize,noabbrev]{cleveref}
\usepackage{algorithm}
\usepackage{algorithmic}
\usepackage{todonotes}[tiny]

\usepackage{kotex}
\usepackage{xcolor}
\definecolor{mycite}{HTML}{CD595D}
\hypersetup{
    colorlinks=true,
    linkcolor=mycite,
    citecolor=mycite,
    urlcolor=mycite,
    filecolor=mycite
}

\title{Score-Based One-step MeanFlow Policy Optimization}

\author{
  \textbf{Kyungyoon Kim}\textsuperscript{1} \quad
  \textbf{Donghyeon Ki}\textsuperscript{1} \quad
  \textbf{Hee-Jun Ahn}\textsuperscript{1} \quad
  \textbf{Byung-Jun Lee}\textsuperscript{1,2} \\[0.5em]
  \textsuperscript{1} Korea University, Decision Making Lab \quad
  \textsuperscript{2} Gauss Labs Inc. \\
  {\tt\small \{kykim803, peop1e1n, niwniwniw, byungjunlee\}@korea.ac.kr}
}

\begin{document}

\maketitle

\begin{abstract}
Diffusion and flow matching have emerged as expressive policy classes in reinforcement learning, but their reliance on multi-step denoising imposes substantial computational overhead at inference time, which is particularly problematic in online RL. MeanFlow offers a promising alternative by learning an average velocity field that maps noise to data in a single network evaluation. However, MeanFlow typically requires samples from the target distribution to construct its target velocity field, which are unavailable in online RL. We propose Score-Based One-step MeanFlow Policy Optimization (SOM), an actor-critic algorithm that resolves this by constructing the target velocity field directly from the Q-function via score estimation and a probability flow ODE, thereby concentrating probability mass on high-value modes. In the fully online RL setting, SOM achieves state-of-the-art performance on locomotion tasks with a single generation step, while substantially reducing both training and inference time compared to prior diffusion- and flow-matching-based policies.
\end{abstract}

\section{Introduction}
Reinforcement learning in continuous action spaces has long been dominated by Gaussian policies, yet their unimodal nature fundamentally limits expressivity, failing to capture the complex, multimodal structure that arises in many real-world control tasks.
To overcome this, recent works have adopted generative models as policy parameterization. 
In particular, diffusion-based generative modeling~\citep{song2019generative, ho2020denoising, song2020denoising, song2021scorebased} and flow matching~\citep{lipman2023flow, liu2023flow} have emerged as promising alternatives to Gaussian policies in RL~\citep{ajay2022conditional, janner2022planning, yang2023policy, wang2023diffusion, ding2024diffusion, chen2024score, wang2024diffusion, psenka2024learning, ren2024diffusion, ki2025prior, chi2025diffusion, ma2025efficient}.
These approaches leverage the expressivity of generative models to represent complex, multimodal action distributions, enabling richer exploration and capturing diverse optimal behaviors.

Despite these advantages, generative policies share a critical limitation: they require multiple, iterative network evaluations to generate a single action. This introduces significant computational overhead at inference time, which is particularly problematic in online RL, where policy inference is continuously interleaved with data collection. Numerous efforts have been made to alleviate the iterative nature of diffusion and flow-matching models, including diffusion distillation~\citep{salimans2022progressive, yin2024one}, consistency models~\citep{song2023consistency, lu2025simplifying}, rectified flow~\citep{liu2023flow}, and shortcut models~\citep{frans2025one}. 

More recently, MeanFlow~\citep{geng2026mean} has been proposed as a promising solution to the multi-step generation problem.
Whereas flow matching learns an instantaneous velocity field that must be integrated over multiple discretization steps, MeanFlow learns a mean velocity field that directly maps noise to data across any time interval in a single network evaluation.
This formulation enables high-quality sample generation in a single step, significantly reducing the computational cost of sampling.

However, directly applying MeanFlow to online RL is not straightforward.
The MeanFlow training objective requires a target velocity field defining the ground-truth transport from noise to the target data distribution, which in turn requires ground-truth samples from that distribution.
In online RL, where such samples are unavailable, constructing an appropriate target velocity without access to explicit target-distribution samples is a key requirement.

In this work, we propose \textbf{Score-Based One-step MeanFlow Policy Optimization (SOM)}, which addresses this challenge by constructing a target velocity field directly from the Q-function, without requiring samples from the target distribution.
Treating the negative Q-function as the energy of the target distribution, we reduce velocity field construction to a score estimation problem.
We adapt the Monte Carlo score estimator of iDEM~\citep{akhound2024iterated}, originally proposed for energy-defined target distributions, to the online RL setting.

Our central contribution is to show that this Q-derived score estimator can be incorporated into a score-based probability flow ODE (PF-ODE), yielding a novel target velocity field.
The resulting velocity field transports noise to the high-value action distribution induced by the current Q-function while maintaining the diversity needed for exploration.
This allows MeanFlow policies to be trained in fully online RL where target-distribution samples are unavailable.

Our main contributions are as follows:
\begin{itemize}
    \item We identify the absence of target-distribution samples as a key bottleneck in applying MeanFlow to online RL, and address it by constructing a target velocity field directly from Q-function gradients.
    \item Building on this insight, we propose Score-Based One-step MeanFlow Policy Optimization (SOM), an actor-critic algorithm that integrates MeanFlow-based generative policies with online RL. Unlike prior diffusion and flow matching policies that require iterative denoising at inference time, SOM achieves expressive policy learning with single-step action sampling.
    \item We demonstrate that SOM achieves state-of-the-art performance on locomotion tasks with just a single generation step, while substantially reducing training and inference time compared to prior diffusion- and flow-matching-based policies.
\end{itemize}
\begin{figure}[t!]
    \centering
    \includegraphics[width=1.0\linewidth]{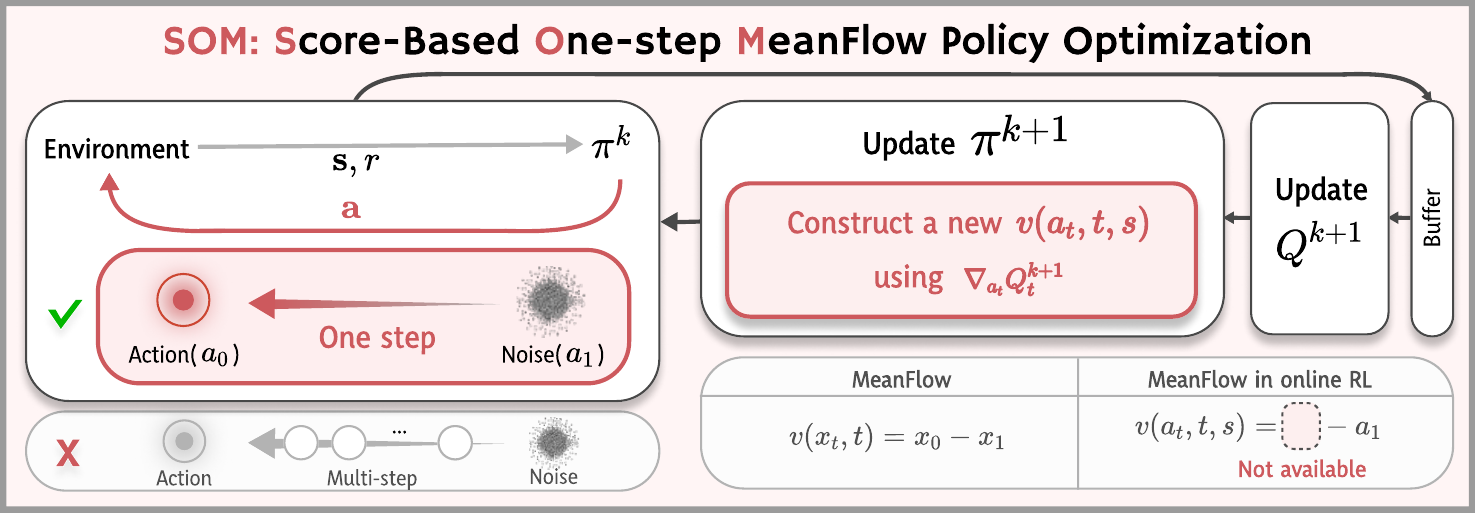}
    \vspace{-0.1in}
    \caption{Unlike existing diffusion and flow matching algorithms, SOM does not require multi-step denoising and can perform denoising in just one step. Furthermore, to train MeanFlow in online RL, SOM directly constructs a target velocity from the Q-function without requiring target-distribution samples.}
    \label{fig:figure1}
\end{figure}
\section{Related Work}
\label{gen_inst}
Recent works have explored diffusion and flow-based models as policy representations in reinforcement learning, leveraging their expressiveness and multimodality to overcome the limitations of conventional unimodal policies. 
QVPO~\citep{ding2024diffusion} introduces a Q-weighted variational lower bound objective that serves as a tight lower bound of the online RL policy objective, enabling diffusion policies to be trained without optimal action samples. 
QSM~\citep{psenka2024learning} regresses the policy's score onto the Q-function's action gradient. 
SDAC and DPMD~\citep{ma2025efficient} reweight the denoising score matching loss with the exponential of the Q-function to approximate the Boltzmann policy induced by the Q-function. They differ in their sampling strategy: DPMD samples from the current policy, while SDAC samples from a Gaussian conditioned on noisy actions. 
DACER~\citep{wang2024diffusion} treats the reverse diffusion process as the policy and backpropagates Q-function gradients across all denoising steps to maximize expected Q-values, while estimating policy entropy via a Gaussian mixture approximation to adaptively regulate exploration. 
DIPO~\citep{yang2023policy} updates replay buffer actions along the Q-function gradient and fits the improved actions with a diffusion model. 

Despite their progress, all of these methods rely on iterative multi-step denoising at inference, which fundamentally limits their efficiency and scalability for online RL where actions must be sampled at every environment step. A recent work, MFP~\citep{zhan2026mean} extends MeanFlow — a one-step generative framework based on average velocity fields — to offline-to-online RL by constructing target velocities from replay buffer samples.
Building on the recent success of MFP~\citep{zhan2026mean} in applying MeanFlow to RL, SOM derives the target velocity field directly from Q-function gradients via a score-based formulation over an analytically tractable Boltzmann distribution, enabling stable training.

\section{Background}
\label{headings}
\subsection{Reinforcement Learning}
\label{background:rl}
Reinforcement learning (RL) problems are typically formalized as a Markov Decision Process, defined by the tuple $\mathcal{M} = \langle \mathcal{S}, \mathcal{A}, \mathcal{P}, r, \rho_0, \gamma \rangle$~\citep{sutton1998reinforcement}. Here, $\mathcal{S} \subseteq \mathbb{R}^n$ denotes the state space, $\mathcal{A} \subseteq \mathbb{R}^m$ the action space, $\mathcal{P}(s'|s,a)$ the state transition function specifying the probability of transitioning to state $s'$ after taking action $a$ in state $s$, $r: \mathcal{S} \times \mathcal{A} \to [r_{\min}, r_{\max}]$ the bounded reward function, $\rho_0 \in \Delta(\mathcal{S})$ the initial state distribution, and $\gamma \in [0,1)$ the discount factor.
The goal of RL is to learn a policy $\pi^*$ that maximizes the expected cumulative discounted return: $J(\pi) = \mathbb{E}_{\tau \sim \pi}\left[ \sum_{t=0}^{\infty} \gamma^t r(s_t, a_t) \right]$.
To this end, the action-value function $Q^\pi(s,a) = \mathbb{E}_{\tau \sim \pi}[\sum_{t=0}^{\infty} \gamma^t r(s_t, a_t) \mid s_0=s, a_0=a]$ measures the expected return of taking action $a$ in state $s$ and following $\pi$ thereafter. Following entropy-regularized RL, we consider the corresponding Boltzmann-form optimal policy: $\pi^*(a \mid s) \propto \exp\left(\alpha Q(s,a)\right)$ where $\alpha$ is a temperature parameter.

\subsection{Variance Preserving SDE}
\label{background:vp_sde}
The forward SDE describes the gradual perturbation of clean data into noise over time. Depending on the choice of the drift term $f(t)$ and the diffusion term $g(t)$, the forward SDE admits several variants, among which the two most common are the Variance Exploding (VE) and Variance Preserving (VP) SDEs. We primarily focus on the VP-SDE, defined as
\begin{align}
    dx_t = -\frac{1}{2}\beta(t)\,x_t\,dt + \sqrt{\beta(t)}\,dw_t, \nonumber
\end{align}
which corresponds to $f(t) = -\frac{1}{2}\beta(t)$ and $g(t) = \sqrt{\beta(t)}$, where $\beta(t)$ denotes a time-dependent noise schedule. Under this SDE, the forward perturbation kernel admits the closed-form Gaussian transition
\begin{align}
\label{eq:vp_sde_forward_kernel}
    p_t(x_t \mid x_0) = \mathcal{N}\!\left(x_t;\, x_0\, e^{-\frac{1}{2}\int^t_0 \beta(\tau)d\tau},\; \bigl(1 - e^{-\int^t_0 \beta(\tau)d\tau}\bigr)\mathbf{I}\right),
\end{align}
so that $x_t$ can be obtained directly from $x_0$ via reparameterization, bypassing explicit SDE simulation.
A description of the VE SDE is provided in the Appendix~\ref{app:ve_sde}.

\subsection{Probability Flow ODE}
\label{background:pf_ode}
For a general SDE $dx_t=f(t)x_tdt + g(t)dw$, where $f(t):\mathbb{R}^D \rightarrow \mathbb{R}^D$ denotes the deterministic drift function, $g(t) \in \mathbb{R}$ is a scalar diffusion coefficient controlling the magnitude of stochastic perturbations, and $w(t)$ is a standard Wiener process (Brownian motion). Since our goal is to construct a deterministic velocity field that transports samples toward high-value regions, the probability flow ODE provides a principled way to derive such a velocity directly from the evolving marginal density.
The corresponding probability flow ODE is given by:
\begin{align}
\label{eq:pf-ode}
    v_t(x_t) := f(t)x_t - \frac{1}{2}g^2(t)\nabla_{x_t}\log p_t(x_t)
\end{align}
This ODE shares the same marginal densities $p_t(x_t)$ as the original SDE.
The equivalence of marginal distributions is guaranteed by the Fokker–Planck equation, which characterizes the time evolution of densities induced by the SDE and its corresponding probability flow ODE.

\subsection{MeanFlow}
\label{background:meanflow}
The probability flow ODE defines an instantaneous velocity at each time $t$, requiring numerical integration over many steps to generate samples.
MeanFlow~\citep{geng2026mean} is a one-step generative modeling framework that introduces a new paradigm, the \emph{average velocity} over a certain time interval, in contrast to the instantaneous velocity modeled by Flow Matching.
Following the Flow Matching framework, given data $x_0 \sim p_{\text{data}}$ and prior $\epsilon \sim p_{\text{prior}}$, we construct a conditional flow path $x_t = a_t x + b_t \epsilon$ with predefined schedules $a_t$ and $b_t$; under the standard choice $a_t = 1-t$, $b_t = t$, the time 
derivative yields the \emph{conditional velocity} $v_t(x_t \mid x_0, \epsilon) = \epsilon - x_0$.
Since the same point $x_t$ may originate from many different $(x_0, \epsilon)$ pairs, Flow Matching takes as its ground-truth target the \emph{marginal velocity field}, $v(x_t, t) \triangleq \mathbb{E}\!\left[\,v_t \,\middle|\, x_t\,\right]$, obtained by taking expectation over all conditional velocities that pass through $x_t$.
Building on this marginal field, MeanFlow introduces the \emph{average velocity} $u(x_t, r, t)$, defined as the displacement between two time steps $r$ and $t$ divided by the elapsed interval $u(x_t, r, t) \triangleq \frac{1}{t-r} \int_{r}^{t} v(x_\tau, \tau)\, d\tau$.
Differentiating both sides of $(t-r)u(x_t, r, t) = \int_{r}^{t} v(x_\tau, \tau)\, d\tau$ with respect to $t$ yields the MeanFlow Identity~\citep{geng2026mean}: 
\begin{align*}
    u(x_t, r, t) = v(x_t, t) - (t-r)\frac{d}{dt}u(x_t, r, t),
\end{align*}
where the total derivative $\tfrac{d}{dt}u = v(x_t, t)\partial_{x_t} u + \partial_t u$ can be efficiently computed via a Jacobian-vector product (JVP). 
A neural network $u_\theta(x_t, r, t)$ is then trained to satisfy this identity by minimizing
\begin{align}
\label{eq:meanflow_loss}
    \mathcal{L}(\theta) = \mathbb{E}_{t,r,x_t}\,\| u_\theta(x_t, r, t) - \mathrm{sg}(u_{\text{tgt}}) \|_2^2,
\end{align}
where $u_{\text{tgt}} = v(x_t, t) - (t-r)\big(v(x_t, t)\,\partial_{x_t} u_\theta + \partial_t u_\theta\big)$ and $\mathrm{sg}(\cdot)$ denotes the stop-gradient operator. 
Following the conditional Flow Matching loss~\citep{lipman2023flow}, the $v(x_t, t)$ in $u_{\text{tgt}}$ is replaced by the conditional velocity $v_t(x_t \mid x_0,\epsilon)$ during training.
At inference time, a sample is generated in a single network evaluation via $x_0 = x_1 - u(x_1, 0, 1)$, where $x_1 = \epsilon \sim p_{\text{prior}}(\epsilon)$.

\subsection{Iterated Denoising Energy Matching}
\label{background:idem}
iDEM~\citep{akhound2024iterated} addresses the setting where the target distribution is defined implicitly through an energy function, $p_0(x) \propto \exp(-E(x))$, with access only to $E(x)$ and its gradient $\nabla E(x)$ rather than samples.
Since samples from $p_0$ are unavailable, the posterior expectation $\nabla_{x_t} \log p_t(x_t) = \mathbb{E}_{x_0 \sim p(x_0|x_t)}[\nabla_{x_t} \log p_t(x_t|x_0)]$ cannot be evaluated directly. Instead, iDEM approximates it via self-normalized importance sampling. Under the VE SDE, samples $x_{0 \mid t}^{(i)}$ are drawn from a Gaussian proposal $\mathcal{N}(x_t, \sigma^2_t \mathbf{I})$ and reweighted by $w_i \propto \exp(-E(x^{(i)}_{0|t}))$ to correct for the proposal--posterior mismatch:
\begin{equation}
\label{eq:idem}
    \nabla_{x_t} \log p_t(x_t) \approx \frac{\frac{1}{K}\sum_i \nabla_{x_t} \exp(-E(x^{(i)}_{0|t}))}{\frac{1}{K}\sum_i \exp(-E(x^{(i)}_{0|t}))}
    = \nabla_{x_t} \log \sum_i \exp(-E(x^{(i)}_{0|t})),
\end{equation}
where $x^{(1)}_{0|t}, \dots, x^{(K)}_{0|t} \sim \mathcal{N}(x_t, \sigma^2_t \mathbf{I})$. Specifically, $x^{(i)}_{0|t}$ is obtained from $x_t$ through the reparameterization $x^{(i)}_{0|t} = x_t + \epsilon^{(i)}$ with $\epsilon^{(i)} \sim \mathcal{N}(0, \sigma^2_t \mathbf{I})$. The notation $0|t$ thus represents a sample drawn at the base time level $t=0$ from a Gaussian distribution whose mean is $x_t$. Moreover, $x_t$ need not be sampled from a distribution conditioned on $x_0$; instead, it can be regarded as an arbitrary fixed point.

\section{Score-Based One-step MeanFlow Policy Optimization}
\label{others}
In this section, we propose a score-based MeanFlow training framework for online RL that obviates the need for sampling from the target distribution.
We begin by introducing a time-dependent energy function that defines a Boltzmann distribution at each timestep in \cref{sec:time_dependent_energy_function}.  
We then present a critic-based method to estimate its score without access to target samples in \cref{sec:estimation_of_time_dependent_energy_gradient}.  
Finally, we convert the estimated score into a velocity field via the PF-ODE formulation in \cref{sec:vel_score}.  

\subsection{Time-Dependent Energy Function}
\label{sec:time_dependent_energy_function}
MeanFlow objectives (\cref{eq:meanflow_loss}) regress $u_\theta(x_t,r,t)$ toward a target $u_{tgt}$ defined by a marginal velocity field. However, their inherent reliance on explicit samples from the target distribution poses a key challenge in online reinforcement learning. Unlike supervised settings, online RL lacks a fixed, externally specified target distribution, making MeanFlow challenging to apply directly in this setting, as it requires samples from the target distribution for training.
Recent approaches such as MFP circumvent this issue by employing Best-of-$N$ (BoN) samples, where $N$ candidate actions are sampled from the current policy and the one with the highest learned $Q$-value is retained. 
While this strategy successfully bypasses the need for explicit value-maximizing target distribution samples, it may struggle as action dimensionality grows, since the number of candidates required to adequately cover the action space can grow exponentially with dimension. 

To overcome the aforementioned limitations, we present a method for constructing the target velocity via gradient information, which scales more favorably with action dimensionality than BoN-based approaches.
In particular, under a Boltzmann-form policy parameterization motivated by entropy-regularized RL, the target score can be naturally expressed in terms of the critic.
To this end, we construct a time-dependent family of distributions that provides a tractable score at each timestep.
Let $p_0(a_0 \mid s)$ denote the target distribution we aim to sample from, which we assume admits an energy-based representation:
\begin{align}
    p_0(a_0 \mid s) = \frac{\exp(\alpha Q(s, a_0))}{Z(s)}, \nonumber
\end{align}
where $Z(s)=\int \exp(\alpha Q(s, a_0))da_0$ is the normalization constant, and $\alpha > 0$ is a temperature parameter. 
The score of this distribution is available in closed form through the learned critic: $\nabla_{a_0} \log p_0(a_0 \mid s) = \alpha \nabla_{a_0} Q(s, a_0)$.
Since PF-ODE-based training requires the score at every timestep $t \in [0,1]$, we construct a time-dependent energy function that preserves the Boltzmann structure of the target across timesteps and admits a tractable score throughout.

Under a forward noising kernel $p_t(a_t|a_0)$, which gradually corrupts a clean sample $a_0$ into Gaussian noise as $t$ increases, the induced density at time $t$ is
\begin{align*}
  p_t(a_t \mid s) &= \int p_t(a_t|a_0)p_0(a_0 \mid s)da_0 = \frac{1}{Z(s)}\int p_t(a_t|a_0)\exp (\alpha Q(s, a_0))da_0
\end{align*}
We additionally define the time-dependent energy function $Q_t(s,a_t)$ implicitly as: 
\begin{align}
  Q_t(s, a_t) = \log \int p_t(a_t|a_0) \exp (\alpha Q(s, a_0))da_0 \nonumber
\end{align}
Under this definition, the density at time $t$ admits an energy-based representation $p_t(a_t \mid s) = \frac{\exp (Q_t(s, a_t))}{Z(s)}$, and the score reduces to the simple form $\nabla_{a_t} \log p_t(a_t \mid s) = \nabla_{a_t} Q_t(s, a_t)$. 
However, computing $Q_t(s, a_t)$ in closed form is generally intractable, as it requires computing the log-expectation of exponentiated Q-values under the forward noising kernel.
Crucially, however, our objective only requires the gradient $\nabla_{a_t} Q_t(s,a_t)$, allowing us to directly approximate it without ever evaluating the intractable $Q_t(s, a_t)$, which is sufficient to construct the probability flow velocity field.

\subsection{Estimation of the Time-Dependent Energy Gradient}
\label{sec:estimation_of_time_dependent_energy_gradient}
Under the Boltzmann action distribution $\pi(\cdot \mid s) \propto \exp(\alpha Q(s, \cdot))$, we consider an energy-based formulation with $E(s,a) = -\alpha Q(s, a)$, where $\alpha > 0$ controls the exploration–exploitation trade-off.
Starting from the iDEM estimator, we substitute the energy function using the Boltzmann-form action distribution introduced above and adapt the resulting estimator to the VP-SDE setting in \cref{background:vp_sde}. Applying these modifications to \cref{eq:idem} yields the following score estimator:
\begin{align}
\label{eq:nablaQt}
    \nabla_{a_t}Q_t(s, a_t) &\approx \frac{\frac{1}{K}\Sigma_i \nabla_{a_t} \exp(\alpha Q(s, a^{(i)}_{0|t}))}{\frac{1}{K}\Sigma_i \exp(\alpha  Q(s,a^{(i)}_{0|t}))} = \nabla_{a_t} \log\Sigma_i \exp(\alpha Q(s,a^{(i)}_{0|t})),
\end{align}
where $a^{(1)}_{0|t}, \dots, a^{(K)}_{0|t} \sim \mathcal N\left(\frac{a_t}{e^{-\frac12\int_0^t \beta(r)\,dr}}, \frac{1-e^{-\int_0^t \beta(r)\,dr}}{e^{-\int_0^t \beta(r)\,dr}}\mathbf{I}\right)$.
This estimator corresponds to the gradient of a locally Gaussian-smoothed Q-landscape, effectively biasing action updates toward high-value regions while maintaining exploration.

\subsection{Velocity Field and Its Relation to the Score Function}
\label{sec:vel_score}
The Probability Flow ODE (PF-ODE) in \cref{background:pf_ode} establishes a connection between the velocity field and the gradient of the time-dependent energy function. 
In the reinforcement learning setting, the action distribution is conditioned on the state $s$. Substituting the VP-SDE coefficients $f(t)$ and $g(t)$ from \cref{background:vp_sde} into the PF-ODE therefore yields the following state-conditioned VP-specific velocity field $v_t(a_t \mid s) = -\tfrac{1}{2}\beta(t)\bigl(a_t + \nabla_{a_t}\log p_t(a_t \mid s)\bigr)$. 
Recall from \cref{sec:time_dependent_energy_function} that the density at timesteps $t$ admits the energy-based form $p_t(a_t \mid s) = \frac{\exp (Q_t(s, a_t))}{Z(s)}$, so that the score reduces to $\nabla_{a_t} \log p_t(a_t \mid s) = \nabla_{a_t} Q_t(s, a_t)$. Specifically, the target velocity is substituted with
\begin{align}
\label{eq:newtargetvel}
    v^{new}_t(a_t\mid s)=-\tfrac{1}{2}\beta(t)(a_t + \nabla_{a_t}Q_t(s, a_t)),
\end{align}
so that the transport velocity is directly determined by the gradient of the time-dependent energy function.
Substituting the PF-ODE–induced velocity field into the MeanFlow objective, where $u_\theta$ denotes the MeanFlow policy parameterization, we obtain the following training formulation.
\begin{align*}
    u^{new}_{tgt} = v^{new}_t(a_t \mid s)-(t-r)(v^{new}_t(a_t \mid s)\partial_{a_t} u_\theta + \partial_t u_\theta)
\end{align*}
The training objective is therefore defined as
\begin{align}
\label{eq:newobj}
    \mathcal{L}(\theta)=\mathbb{E}_{(s,a_0)\sim \mathcal{D},\, t\sim \mathcal{U}(0,1),\, a_t\sim p_t(a_t|a_0)}\left[||u_\theta(a_t,r,t,s)-sg(u^{new}_{tgt})||^2_2\right].
\end{align}

\subsection{Implementation Details}
\paragraph{Normalization and Re-scaling.}
We apply two normalizations to stabilize training. 
First, we found that the highly varying scale of the score $\nabla_{a_t} Q_t(s, a_t)$ can result in severely imbalanced updates across states and timesteps, which can slow down learning;
we normalize the score to have unit norm and rescale it with a tunable coefficient $w$. 
Accordingly, \cref{eq:newtargetvel} can be rewritten as:
\begin{equation}
    v^{new}_t(a_t\mid s)=-\tfrac{1}{2}\beta(t)\Big(a_t + w\frac{\nabla_{a_t}Q_t(s, a_t)}{\left\|\nabla_{a_t}Q_t(s, a_t)\right\|_2 + \epsilon}\Big).
\end{equation}

Second, similar to the batch-wise advantage normalization used in GRPO~\citep{shao2024deepseekmath}, we normalize $Q$ values across each training batch when constructing the policy targets. This makes training more robust to varying reward scales across tasks, and reduces the variance of the score estimator.
\paragraph{Critic Update.}
We train the critic via a distributional Bellman objective with twin Q-networks $Q_{\phi_1}, Q_{\phi_2}$ and corresponding target networks following DACER~\citep{wang2024diffusion}.
\paragraph{Best-of-$N$.}
We also incorporate BoN sampling into SOM when training the critic, which makes the critic estimate the action value of further improved policy.
While our score-based target velocity effectively incorporates critic gradient information into actor updates, we find that BoN sampling additionally speeds up training, as scores can sometimes plateau in complex Q landscapes.
Specifically, multiple candidate actions are generated for each state and evaluated using the critic, and the action with the highest predicted Q-value is selected when evaluating the target value in the critic objective.
The overall procedure is described in \cref{alg:alg1}, and the full pipeline is illustrated in \cref{fig:figure1}. 

\section{Experiments}
\label{sec:5}
\begin{figure}[t!]
    \centering
    \includegraphics[width=1.0\linewidth]{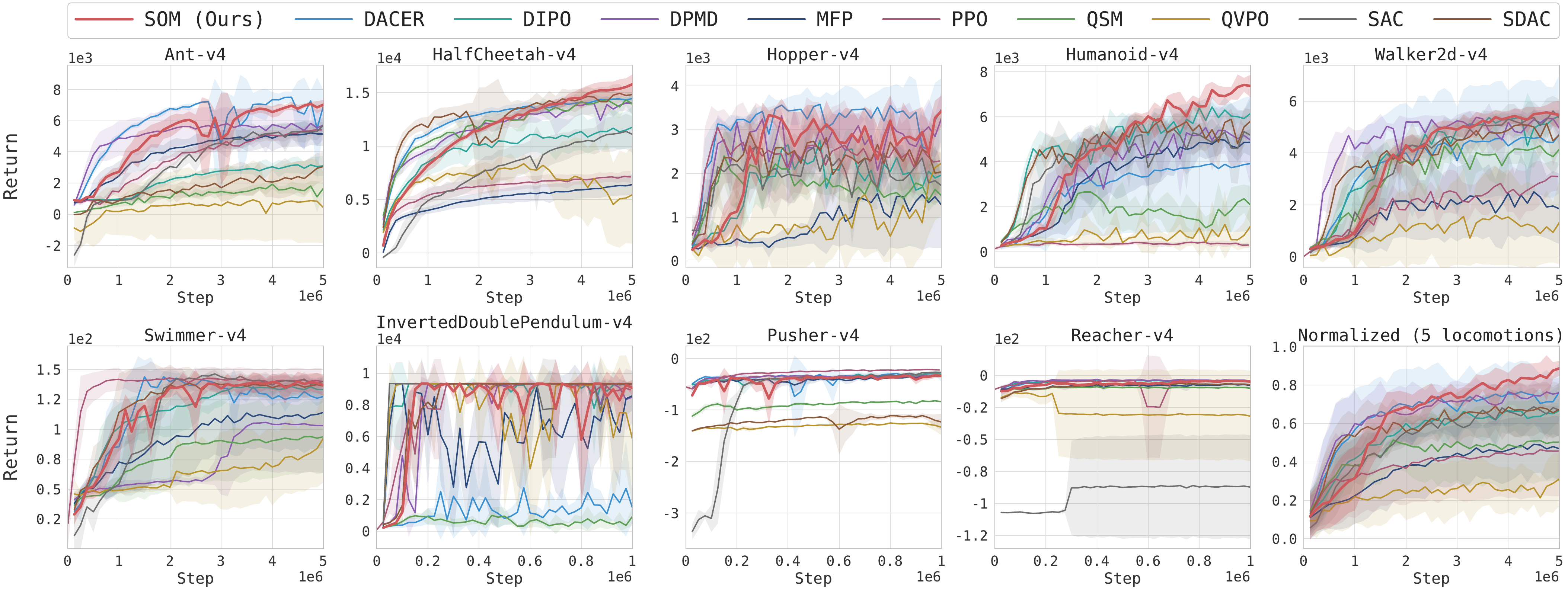}
    \caption{\textbf{Training curves on locomotion benchmarks} Curves denote the mean across five random seeds, with shaded regions representing the 95\% confidence interval. The bottom-right panel reports a per-environment min–max normalized return averaged across the five locomotion benchmarks.}
    \label{fig:figure2}
\end{figure}

\subsection{MuJoCo Benchmarks}
\paragraph{Benchmarks.}
We evaluated the performance on nine tasks from the MuJoCo v4 suite~\citep{todorov2012mujoco, towers2024gymnasium}: \texttt{Ant-v4}, \texttt{HalfCheetah-v4}, \texttt{Hopper-v4}, \texttt{Humanoid-v4}, \texttt{Walker2d-v4}, \texttt{Swimmer-v4}, \texttt{Pusher-v4}, \texttt{Reacher-v4}, and \texttt{InvertedDoublePendulum-v4}. These environments are widely used to benchmark model-free RL algorithms and cover a broad range of state-action dimensionalities. Each algorithm is trained and reported over $5$ random seeds. We employ 1M environment steps for \texttt{InvertedDoublePendulum-v4}, \texttt{Pusher-v4}, and \texttt{Reacher-v4}, and 5M steps for all others.
\paragraph{Baselines.}
We compare our method against a diverse set of baselines covering both generative and non-generative policy classes. Among generative policy baselines, we include diffusion-based actors—DACER~\citep{wang2024diffusion}, DIPO~\citep{yang2023policy}, DPMD~\citep{ma2025efficient}, QSM~\citep{psenka2024learning}, QVPO~\citep{ding2024diffusion}, and SDAC~\citep{ma2025efficient}—as well as MFP~\citep{zhan2026mean}, which builds on the recently proposed MeanFlow framework~\citep{geng2026mean}. For non-generative baselines, we include the standard off-policy and on-policy actor-critic algorithms SAC~\citep{haarnoja2018soft} and PPO~\citep{schulman2017proximal}, which serve as strong and widely adopted baselines for continuous control. 
All baselines follow the same training protocol as described in the Benchmarks section above, except that PPO is trained for 5M environment steps on \texttt{InvertedDoublePendulum-v4}, \texttt{Pusher-v4}, and \texttt{Reacher-v4} ($5\times$ the x-axis scale of other methods).

\begin{table}
  \caption{Final Episode returns: mean $\pm$ 95\% CI over 5 seeds. \textcolor{mycite}{\textbf{Bold}} indicates the highest mean in each column; \underline{underline} indicates the second highest.}
  \label{main_table}
  \centering
  \resizebox{\textwidth}{!}{%
  \begin{tabular}{lcccccc}
    \toprule
    Algorithm & Ant-v4 & HalfCheetah-v4 & Hopper-v4 & Humanoid-v4 & Walker2d-v4 & Normalized \\
    \midrule
    DACER  & \underline{$6953.5 \pm 1029.7$} & $14445.8 \pm 281.5$ & \textcolor{mycite}{$\mathbf{3462.8 \pm 709.2}$} & $3915.5 \pm 3006.2$ & $4723.6 \pm 2276.1$ & \underline{$0.764$} \\
    DIPO   & $3112.0 \pm 640.7$ & $11762.1 \pm 1912.9$ & $1962.7 \pm 158.8$ & \underline{$6139.4 \pm 645.9$} & $5425.8 \pm 181.3$ & $0.667$ \\
    DPMD   & $5714.1 \pm 887.8$ & $14020.6 \pm 946.3$ & $3257.6 \pm 623.3$ & $4968.4 \pm 1991.1$ & $4953.8 \pm 707.1$ & $0.760$ \\
    QSM    & $1671.0 \pm 377.9$ & $13922.1 \pm 1032.4$ & $1493.3 \pm 146.7$ & $2082.9 \pm 842.4$ & $4144.9 \pm 575.3$ & $0.468$ \\
    QVPO   & $430.3 \pm 2277.9$ & $5447.0 \pm 4585.8$ & $2232.1 \pm 1189.3$ & $1127.1 \pm 937.4$ & $1323.3 \pm 1609.7$ & $0.456$ \\
    SDAC   & $3067.0 \pm 926.6$ & \underline{$14820.9 \pm 1049.0$} & $2036.4 \pm 431.5$ & $5803.5 \pm 520.6$ & $4702.1 \pm 627.4$ & $0.503$ \\
    MFP    & $5150.7 \pm 785.6$ & $6400.3 \pm 966.5$ & $1280.3 \pm 981.8$ & $4862.2 \pm 603.1$ & $1849.8 \pm 1356.2$ & $0.310$ \\
    SAC    & $5711.6 \pm 530.4$ & $11098.7 \pm 1697.4$ & $1722.6 \pm 248.7$ & $5179.5 \pm 335.0$ & \underline{$5348.6 \pm 712.9$} & $0.669$ \\
    PPO    & $5417.2 \pm 283.6$ & $7112.6 \pm 576.7$ & $2282.5 \pm 770.3$ & $299.2 \pm 119.0$ & $3096.3 \pm 540.3$ & $0.674$ \\
    \midrule
    \textbf{SOM (Ours)} & \textcolor{mycite}{$\mathbf{7033.5 \pm 304.2}$} & \textcolor{mycite}{$\mathbf{15789.8 \pm 910.9}$} & \underline{$3425.2 \pm 353.6$} & \textcolor{mycite}{$\mathbf{7365.7 \pm 493.3}$} & \textcolor{mycite}{$\mathbf{5472.8 \pm 526.9}$} & \textcolor{mycite}{$\mathbf{0.887}$} \\
    \bottomrule
  \end{tabular}%
  }
\end{table}

\begin{table}[t]
\caption{\textbf{Per-step training and per-call inference time on HalfCheetah-v4 (MuJoCo).} Training uses batch size 256 and inference uses a single observation (batch size 1). $T_d$: number of denoising steps; $N$: number of best-of-$N$ particles. All times are reported in milliseconds as mean $\pm$ 95\% CI over 30 runs. \textcolor{mycite}{\textbf{Bold}} indicates the fastest in each column, and \underline{underline} indicates the second fastest.}
\label{tab:training_inference_time}
\centering
\small
\setlength{\tabcolsep}{4pt}
\begin{tabular}{lrrrr@{\hskip 1em}lrrrr}
\toprule
Method & $T_d$ & $N$ & Training (ms) & Inference (ms) &
Method & $T_d$ & $N$ & Training (ms) & Inference (ms) \\
\midrule
SOM   & 1   & 32 & \textcolor{mycite}{\textbf{6.512 $\pm$ 0.026}} & \textcolor{mycite}{\textbf{0.218 $\pm$ 0.002}} & SDAC  & 20  & 32 & 15.633 $\pm$ 0.057 & 1.012 $\pm$ 0.014 \\
QSM   & 20  & 32 & \underline{7.238 $\pm$ 0.032} & \underline{0.606 $\pm$ 0.003} & DACER & 20  & 1  & 15.791 $\pm$ 0.161 & 0.703 $\pm$ 0.010 \\
DPMD  & 20  & 32 & 8.941 $\pm$ 0.037 & 1.036 $\pm$ 0.004 & QVPO  & 20  & 32 & 40.768 $\pm$ 0.136 & 1.044 $\pm$ 0.027 \\
DIPO  & 100 & 1  & 10.097 $\pm$ 0.042 & 1.562 $\pm$ 0.029 &       &     &    &                    &                   \\
\bottomrule
\end{tabular}
\end{table}

\paragraph{Results.}
\cref{fig:figure2} shows the full training curves over all 9 tasks, and Table~\ref{main_table} reports the final episode returns of all methods across the five locomotion benchmarks, together with the average of per-environment normalized performance. SOM achieves the highest mean return on four out of five environments and the second highest on \texttt{Hopper-v4} where the difference to the top-performing method is statistically insignificant. The improvement is most pronounced on the \texttt{Humanoid-v4} task, which has the highest action dimensionality among all tasks, outperforming the second-best method by a clear margin. This result aligns with our intuition that gradient information becomes increasingly beneficial as action dimensionality grows. 
In addition, \cref{tab:training_inference_time} shows that SOM achieves substantially lower inference latency than all diffusion-based baselines while maintaining efficient training time, highlighting the practical advantage of one-step action generation. 
Additional experimental results under the VE-SDE formulation are provided in \cref{app:ve_sde}. 
Full implementation details, including hyperparameters, are provided in Appendix~\ref{app:implementation}.

\paragraph{Policy Analysis.}
\begin{figure}[t!]
    \centering
    \includegraphics[width=0.90\linewidth]{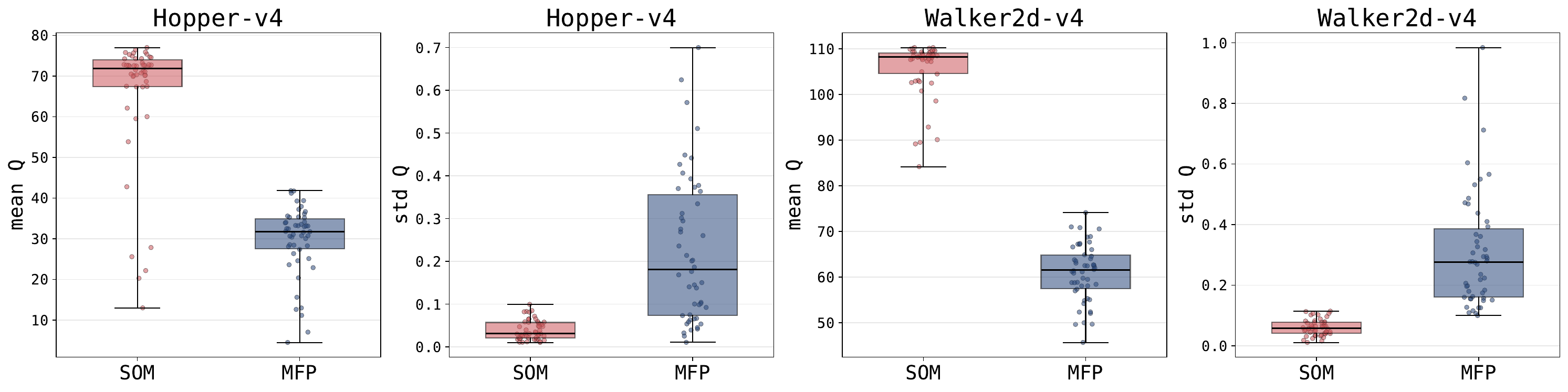}
    \vspace{-4pt}
    \caption{\textbf{$Q$-value mean and std over $N{=}1000$ action samples at $50$ states.} Details in Appendix~\ref{app:bon}.}
    \label{boxplot}
\end{figure}
\begin{figure}[t!]
    \centering
    \includegraphics[width=0.90\linewidth]{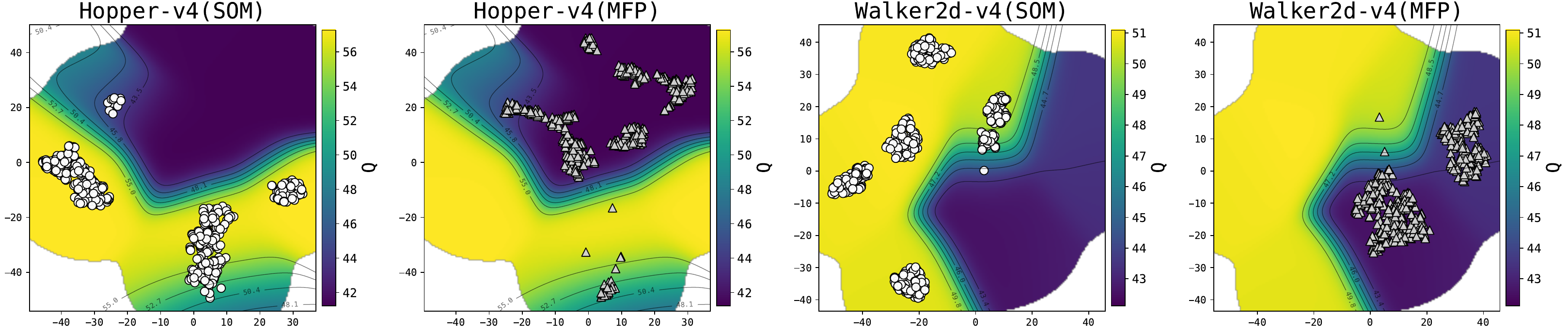}
    \vspace{-4pt}
    \caption{Action samples from SOM (white circles) and MFP (gray triangles) at a fixed state, projected via t-SNE on the $Q$-landscape. Details in Appendix~\ref{app:t-sne}.
}
    \label{fig:tsne}
\end{figure}
SOM can effectively shift the policy distribution toward high-Q regions by incorporating gradient information. Figure~\ref{boxplot} shows that SOM exhibits higher mean Q-values and lower standard deviation compared to MFP, which uses BoN only to improve the policy. This behavior is further supported by the t-SNE visualization in Figure~\ref{fig:tsne}, where SOM samples form tight clusters around high-Q regions, while MFP samples are more diffusely distributed across the action space.

\subsection{SOM Behavior on Bandit Environments}

\begin{figure}[t!]
    \centering
    \includegraphics[width=0.90\linewidth]{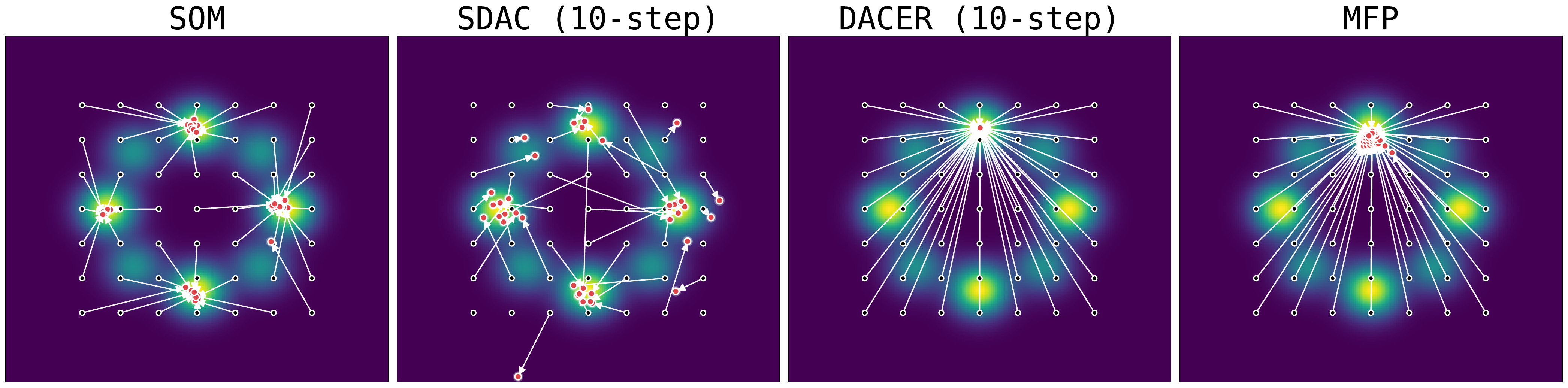}
    \vspace{-0.1in}
    \caption{\textbf{Generative trajectories.}
    Arrow plots from a $7\times 7$ grid of $x_T$ (black) to $x_0$ (red) on the eight-Gaussian reward.
    SDAC and DACER with their full 10-step rollouts.
    SOM (Ours) and MFP, both 1-step by design. Details and additional results in Appendix~\ref{app:2d_task}.
    }
    \label{fig:bandit_main}
\end{figure}
\paragraph{Mode Coverage.}
We verify SOM on a 2D bandit whose reward is a mixture of eight Gaussian modes (Figure~\ref{fig:bandit_main}).
For each method we roll out the learned policy from a fixed $7\times 7$ grid of initial noise $a_1$ and visualize the endpoint $a_0$ along with the arrow connecting $a_1$ to $a_0$.
With their full $10$-step reverse process, the diffusion baselines concentrate samples on the high-reward modes. SDAC distributes its samples relatively evenly across all four high-reward modes, while DACER collapses onto a single mode.
Among one-step samplers, MFP collapses onto a single mode, failing to capture the reward landscape, whereas SOM successfully maps the initial noise grid to all four high-reward modes.
This confirms that SOM learns a coherent one-step map from noise to high-reward actions. We also include experimental results on the two-moons and checkerboard datasets in \ref{app:2d_task}.

\begin{figure}[t!]
    \centering
    \includegraphics[width=0.9\linewidth]{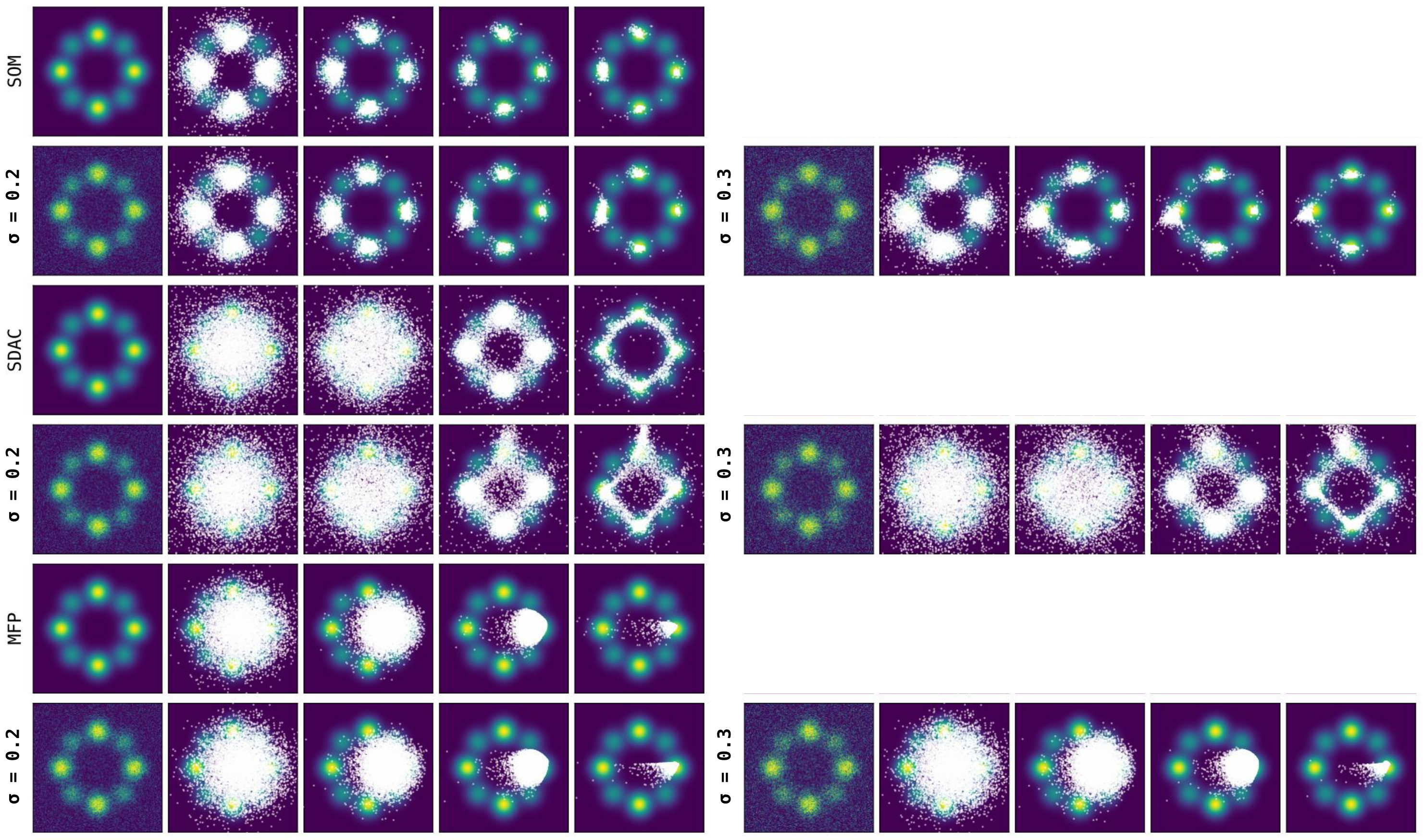}
    \caption{\textbf{Robustness under two reward perturbations.} Each side is a $6\!\times\!5$ grid: rows alternate \emph{Clean} / \emph{Noisy} training Q for SOM, SDAC, and MFP (top to bottom); columns show the Q landscape and the action distribution at sampler times $t\!=\!0.75,0.5,0.25,0.0$. \textbf{(Left)} random Gaussian noise on Q ($\sigma\!=\!0.20$). \textbf{(Right)} random Gaussian noise on Q ($\sigma\!=\!0.30$).
    Details and additional results in Appendix~\ref{app:robustness_q}.
    }
    \label{fig:bandit_noise_q}
\end{figure}

\paragraph{Robustness under Imperfect Critics.}
A central challenge of online RL is that the critic $Q_\phi$ is itself being learned and thus inevitably imperfect, with TD noise and unreliable estimates in under-explored regions. 
The iDEM-style estimator in \cref{eq:nablaQt} mitigates this by returning the gradient of a Gaussian-smoothed Q-landscape, whose smoothing kernel scales with the diffusion noise level $\sigma(t)$: large $t$ yields heavy smoothing that averages out coarse critic errors, while small $t$ preserves local structure near high-value modes.
Fig.~\ref{fig:bandit_noise_q} shows the samples in different timesteps with and without explicit noise in Q-function.
We observe that the smoothed $Q$ and its induced velocity field remain largely preserved under noise—empirically confirming the timestep-adaptive smoothing behavior and offering a potential explanation for SOM's strong empirical performance.

\section{Conclusion}
We presented Score-Based One-step MeanFlow Policy Optimization (SOM), an actor-critic algorithm that trains MeanFlow policies in fully online RL.
The central insight is that the absence of target-distribution samples in online RL, a key challenge for applying MeanFlow in this setting, can be circumvented by constructing the target velocity field directly from the Q-function.
By estimating the score of a Boltzmann target distribution from the Q-function, we incorporate it into a probability flow ODE to obtain a target velocity field that transports noise toward high-value modes.
Empirically, SOM achieves state-of-the-art performance on standard locomotion benchmarks with a single generation step, while substantially reducing both training and inference time compared to prior diffusion-based policies.

\paragraph{Limitations and Future Work.}
SOM has several limitations that suggest promising directions for future work. 
Our evaluation focuses on continuous control from the MuJoCo suite, and extending SOM to higher-dimensional or partially observable settings—such as visual control or robotic manipulation—remains an important next step. 
Finally, SOM's performance is ultimately bounded by the quality of the learned critic; tighter integration with distributional or 
uncertainty-aware critics is left to future work.

\medskip

\bibliographystyle{plain}
\bibliography{example_paper}

\newpage
\appendix

\section{Algorithm Pseudocode}
\label{app:A}
\begin{algorithm}[h!]
\caption{Score-Based One-step MeanFlow Policy Optimization}
\label{alg:alg1}
\begin{algorithmic}
\STATE {\bfseries Function} $\pi(s; u_\theta, Q_\phi, N)$ \hfill $\triangleright$ Best-of-$N$ action selection
\STATE \quad Sample $\epsilon^{(i)} \sim \mathcal{N}(0, \mathbf{I})$ for $i = 1, \dots, N$
\STATE \quad $a^{(i)} \leftarrow \epsilon^{(i)} - u_\theta(\epsilon^{(i)}, 0, 1, s)$ for $i = 1, \dots, N$ \hfill $\triangleright$ One-step generation
\STATE \quad \textbf{return} $a^{(i^*)}$ where $i^* = \arg\max_i Q_\phi(s, a^{(i)})$
\STATE
\STATE {\bfseries Input:} Policy $u_{\theta}$, Critic $Q_\phi$, buffer $\mathcal{B}$, Monte Carlo sample size $K$, BoN size $N$
\FOR{each iteration}
\FOR{each sampling step}
\STATE $a \leftarrow \pi(s; u_\theta, Q_\phi, N)$
\STATE Execute $a$, observe reward $r$ and next state $s'$
\STATE Store transition $(s, a, r, s')$ in buffer $\mathcal{B}$
\ENDFOR
\FOR{each update step}
\STATE Sample mini-batch $\{(s, a, r, s')\} \sim \mathcal{B}$
\STATE Update critic $Q_\phi$ following DACER~\citep{wang2024diffusion}
\STATE \emph{$\triangleright$ Train MeanFlow policy $u_\theta$}
\STATE Sample $t, r \sim \mathcal{U}(0,1)$ with $r \leq t$
\STATE Get $a_t$ from \cref{eq:vp_sde_forward_kernel}
\STATE Get $\nabla_{a_t}Q_t(s, a_t)$ from \cref{eq:nablaQt}
\STATE Get $v^{new}_t(a_t \mid s)$ from \cref{eq:newtargetvel}
\STATE Update $\theta$ to minimize \cref{eq:newobj}
\ENDFOR
\ENDFOR
\STATE \textbf{return} One-step policy $\pi$
\end{algorithmic}
\end{algorithm}

\section{VE-SDE Formulation of SOM and Experimental Results under VE-SDE}
\subsection{Variance Exploding SDE}
\label{app:ve_sde}
\begin{figure}[h!]
    \centering
    \includegraphics[width=1.0\linewidth]{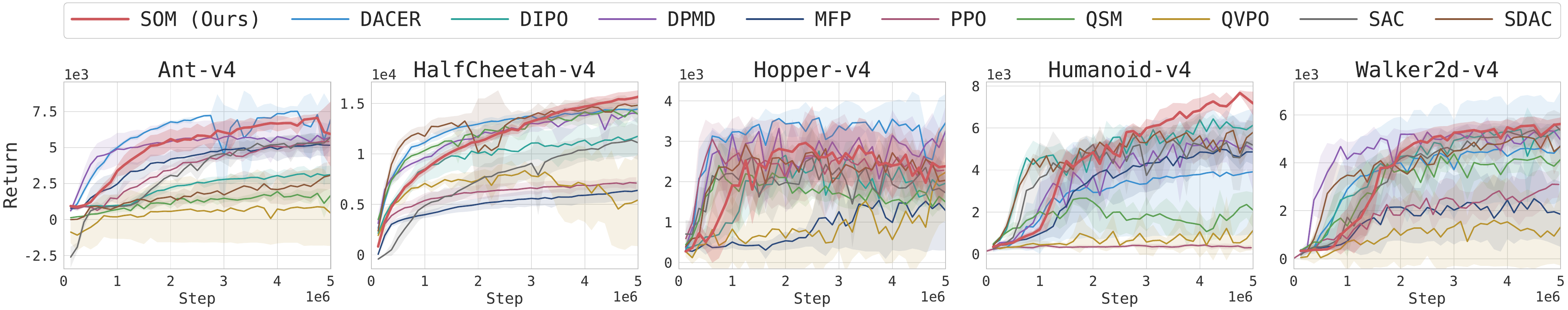}
    \caption{\textbf{VE-SDE Results.}}
    \label{fig:ve_sde}
\end{figure}
The forward SDE describes the process in which clean data is gradually perturbed into noise over time. Based on how the variance evolves over time, the forward SDE can be categorized into three types according to the forms of the drift term $f(t)$ and the diffusion term $g(t)$: the Variance Exploding (VE) SDE, the Variance Preserving (VP) SDE, and the Sub-Variance Preserving (sub-VP) SDE. In this work, we primarily focus on the VE-SDE.

For the Variance Exploding (VE) SDE:
\begin{align}
    dx_t = \sqrt{\frac{d\sigma^2(t)}{dt}}dw
\end{align}
where $\sigma(t)$ is monotonically increasing function over $t \in [0,T]$. The function $\sigma(t)$ controls the noise scale of the diffusion process, satisfying $\sigma^2(T) \gg \sigma^2(0)$, thereby ensuring progressive variance expansion over time.

\subsection{VE-SDE Formulation of SOM}
We have:
\begin{align}
    &f(t)=0, \quad g^2(t) = \frac{d\sigma^2(t)}{dt} \nonumber
\end{align}
Under the VE-SDE, the forward perturbation kernel admits a closed-form Gaussian transition:
\begin{align}
\label{eq:conditionalprob}
    p_t(a_t|a_0)=\mathcal{N}(a_t;a_0,\sigma^2(t)\mathbf{I}).
\end{align}
Thus, perturbed samples can be generated via the reparameterization, which allows efficient sampling without explicitly solving the SDE.
Substituting the VE-SDE coefficients $f(t)$ and $g(t)$ into the PF-ODE therefore yields the following state-conditioned VE-specific velocity field
\begin{align}
    v_t(a_t \mid s)=-\frac{1}{2}\frac{d\sigma^2(t)}{dt}\nabla_{a_t}\log p_t(a_t \mid s). \nonumber
\end{align}
Recall from \cref{sec:time_dependent_energy_function} that the density at timesteps $t$ admits the energy-based form $p_t(a_t \mid s) = \frac{\exp (Q_t(s, a_t))}{Z(s)}$, so that the score reduces to $\nabla_{a_t} \log p_t(a_t \mid s) = \nabla_{a_t} Q_t(s, a_t)$. Specifically, the target velocity is substituted with
\begin{align*}
    v^{new}_t(a_t\mid s)=-\frac{1}{2}\frac{d\sigma^2(t)}{dt}\nabla_{a_t}Q_t(s, a_t). 
\end{align*}

\subsection{Experimental Results under VE-SDE}
We evaluated the performance on nine OpenAI Gym MuJoCo v4 tasks~\citep{todorov2012mujoco}: \texttt{Ant-v4}, \texttt{HalfCheetah-v4}, \texttt{Hopper-v4}, \texttt{Humanoid-v4}, \texttt{Walker2d-v4}, \texttt{Swimmer-v4}, \texttt{InvertedDoublePendulum-v4}, \texttt{Pusher-v4}, and \texttt{Reacher-v4}.

In particular, SOM significantly outperforms existing diffusion-based baselines on the high-dimensional \texttt{Humanoid-v4} task while remaining competitive on the remaining benchmarks. Moreover, consistent with the results in \cref{tab:training_inference_time}, SOM maintains substantially lower training and inference costs than iterative diffusion-based policies, highlighting the computational advantage of one-step action generation under both VP- and VE-SDE formulations.

\section{Ablation Study}
We conduct all ablation experiments on six OpenAI Gym MuJoCo v4 benchmarks~\citep{todorov2012mujoco}: \texttt{Ant-v4}, \texttt{HalfCheetah-v4}, \texttt{Hopper-v4}, \texttt{Humanoid-v4}, \texttt{Swimmer-v4}, and \texttt{Walker2d-v4}.
\subsection{Sensitivity to the Rescaling Coefficient}
\begin{figure}[h!]
    \centering
    \includegraphics[width=1.0\linewidth]{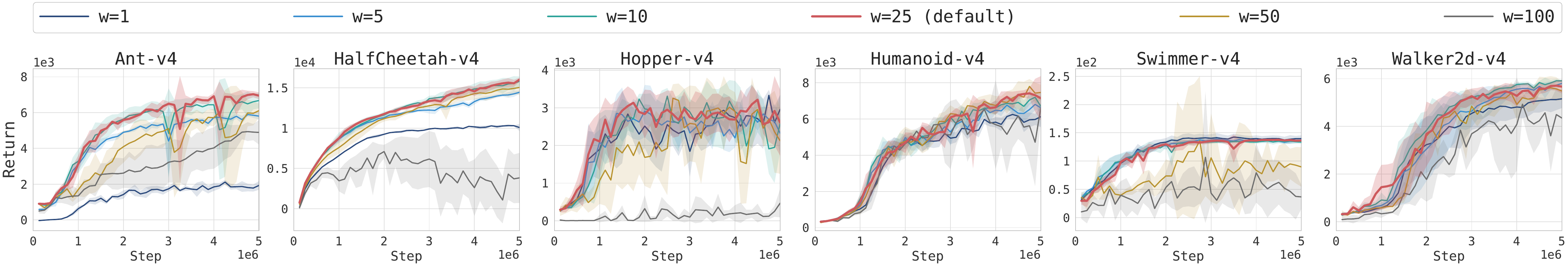}
    \caption{\textbf{Ablation results for the rescaling coefficient $w$.}
    }
    \label{fig:w_ablation}
\end{figure}
As shown in \cref{fig:w_ablation}, moderate values of $w$ generally yield stable and strong performance across environments, whereas excessively small or large values tend to degrade training stability and final return. Small values of $w$ weaken the contribution of the score term in the velocity field, while overly large values can lead to unstable action updates due to excessively strong score guidance. Across most environments, intermediate values achieve the best trade-off between stability and performance. Based on these results, we use $w=25$ as the default setting in all experiments.

\newpage

\subsection{Effect of the Number of Monte Carlo Samples in the iDEM Estimator}

\begin{figure}[h!]
    \centering
    \includegraphics[width=1.0\linewidth]{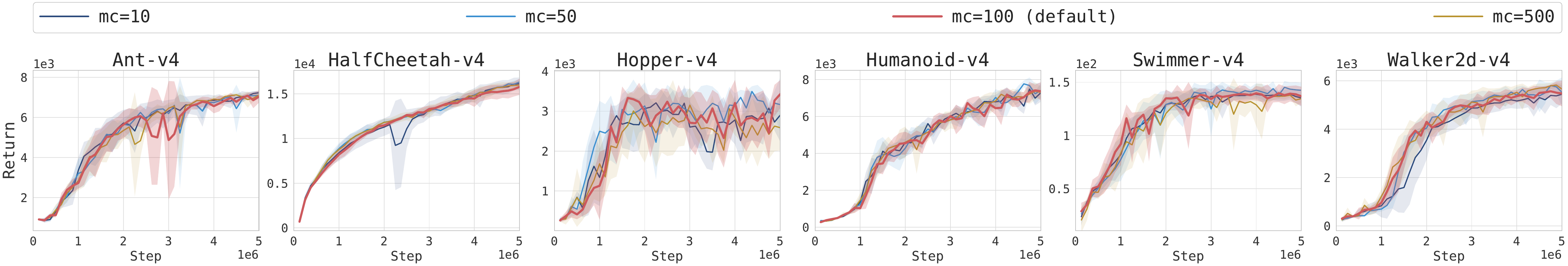}
    \caption{\textbf{\textbf{Ablation results for the number of Monte Carlo samples in the iDEM estimator.}}
    }
    \label{fig:mc_ablation}
\end{figure}

As shown in \cref{fig:mc_ablation}, increasing the number of Monte Carlo samples generally improves training stability and final performance by providing a more accurate approximation of the iDEM score estimator. In particular, small sample sizes exhibit noticeably higher variance and occasional performance collapse in several environments, while larger sample sizes produce more stable learning dynamics. However, the performance gain saturates beyond a certain point, and we find that $N=100$ provides a good trade-off between estimator quality and computational efficiency.

\section{Experimental Setup for the Box Plot}
\label{app:bon}
\paragraph{Batch Construction via Rollout}
We construct an evaluation batch by first generating a fixed-length rollout using the SOM policy as the behavior policy.
Starting from an environment reset, we execute the SOM policy for 800 environment steps, resetting the environment whenever an episode terminates or is truncated.
All visited observations are recorded, yielding a pool of 800 states.
From this pool, we uniformly sample 50 states, which form the evaluation batch.
We conduct this evaluation on the \texttt{Hopper-v4} and \texttt{Walker2d-v4} environments.

\paragraph{Best-of-\texorpdfstring{$N$}{N} Evaluation}
For each sampled state, we evaluate candidate actions generated by two methods: SOM and MFP. Each method generates 32 candidate actions conditioned on the given state. These candidate actions are evaluated using a learned Q-function, producing a set of Q-values per method. We summarize each candidate set using the mean and standard deviation of its Q-values. This results in, for each state, a pair of statistics (mean and standard deviation) for SOM and MFP respectively.

\section{Experimental Setup and Results for the t-SNE Visualization}
\label{app:t-sne}
\subsection{Experimental Setup}
We visualize the action distributions of SOM and MFP policies, both performing one-step inference, at a fixed environment state using t-SNE. For each environment (\texttt{HalfCheetah-v4}, \texttt{Hopper-v4} and \texttt{Walker2d-v4}), we fix an initial state and perform a 5-step rollout using the SOM policy. At each step, we sample 100 actions from both SOM and MFP conditioned on the current observation, resulting in 500 samples per policy.
This design ensures that both policies are evaluated on a shared sequence of observations, so that any differences in the resulting action distributions can be attributed to the policies themselves rather than differences in visited trajectories.

\subsection{Results for the t-SNE Visualization}
\begin{figure}[h!]
    \centering
    \includegraphics[width=1.0\linewidth]{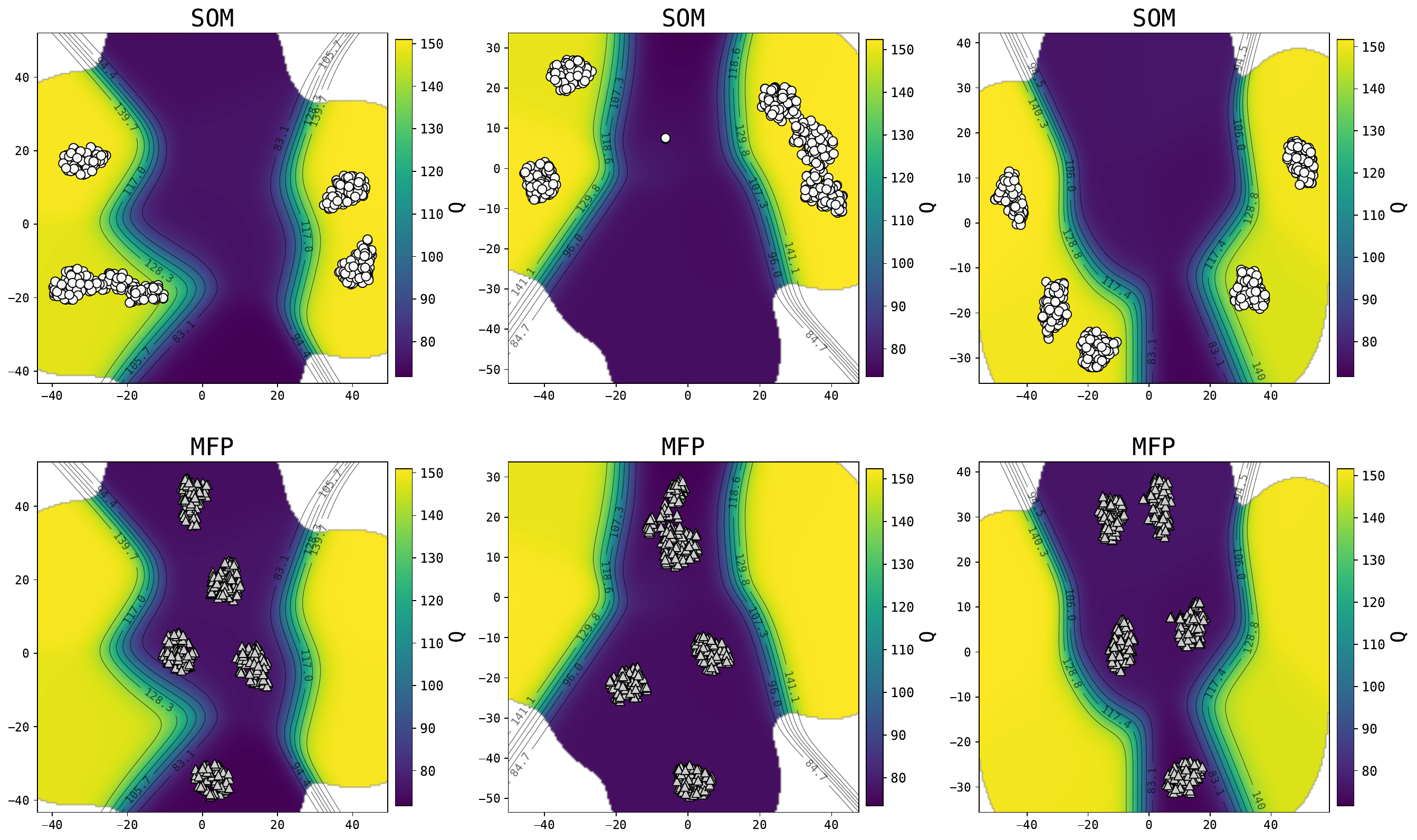}
    \caption{\textbf{HalfCheetah-v4}}
    \label{fig:app_tse_half}
\end{figure}

\begin{figure}[h!]
    \centering
    \includegraphics[width=1.0\linewidth]{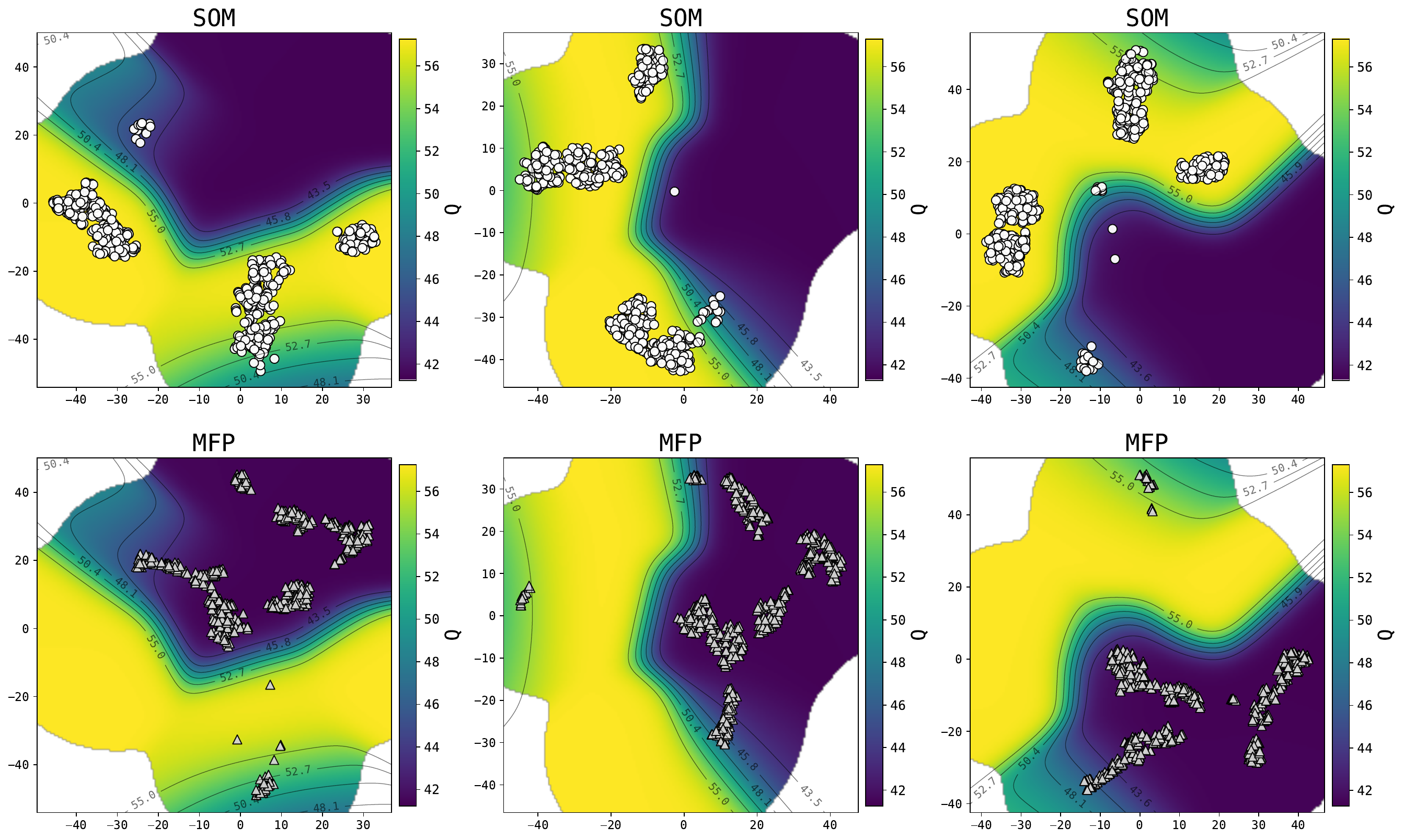}
    \caption{\textbf{Hopper-v4}}
    \label{fig:app_tse_hop}
\end{figure}

\begin{figure}[h!]
    \centering
    \includegraphics[width=1.0\linewidth]{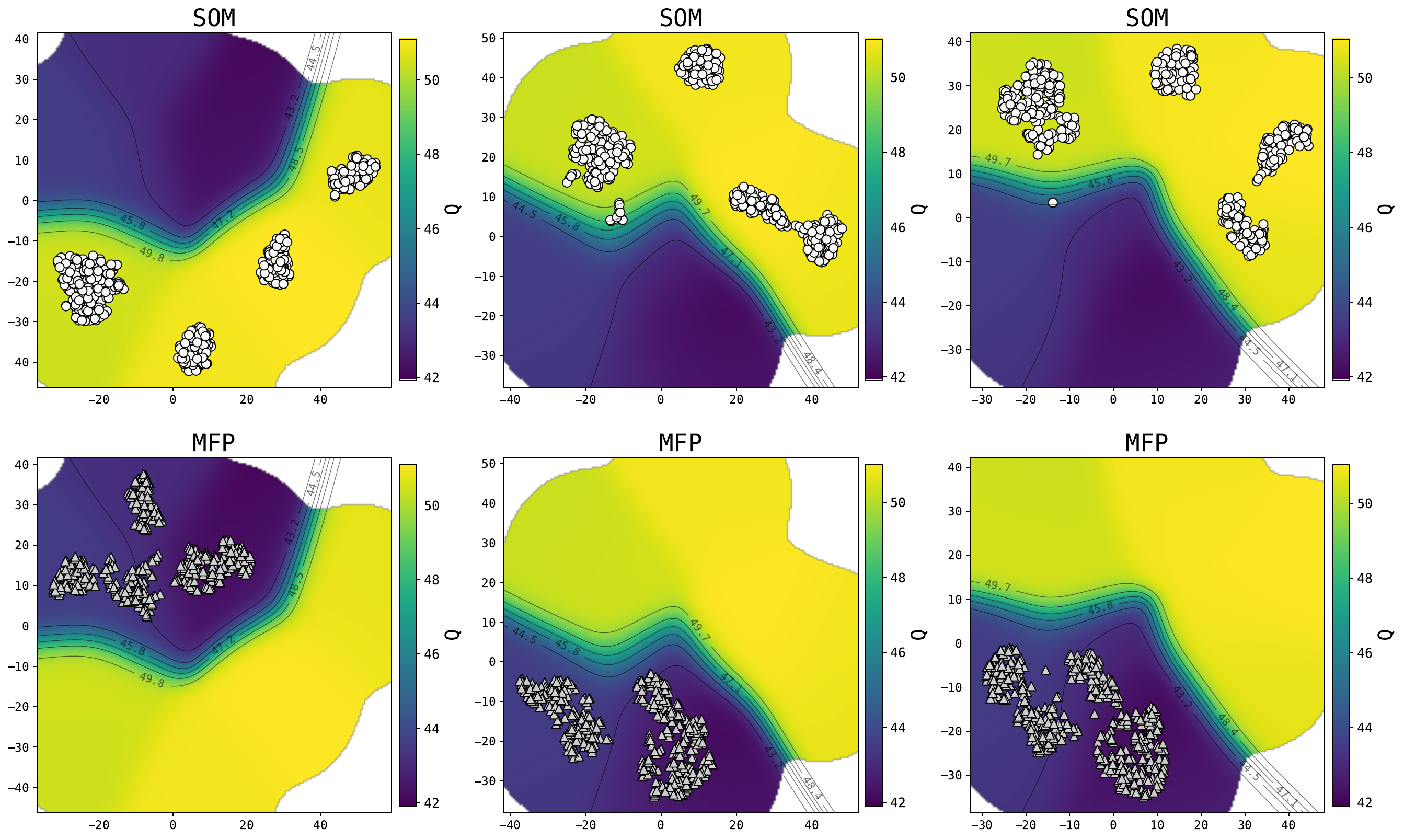}
    \caption{\textbf{Walker2d-v4}}
    \label{fig:app_tse_walk}
\end{figure}

We visualize the resulting Q-landscape on the t-SNE plane using a Gaussian kernel density estimate (KDE) heatmap, with 8 contour lines overlaid for readability. SOM samples are plotted as white circles, and MFP samples as gray triangles.
As shown in \cref{fig:app_tse_half}, \cref{fig:app_tse_hop}, and \cref{fig:app_tse_walk},
SOM forms tighter clusters around Q-optimal modes, while MFP spreads more diffusely.

\section{2D Bandit Tasks: Bandit, Two-Moons, and Checkerboard}
\label{app:2d_task}
\subsection{Eight Mode Bandit}
\paragraph{Reward function}
The reward is defined as a Gaussian-mixture density with $K=8$ isotropic components $\mathcal{N}(\mu_i,\sigma^2 I_2)$ with $\sigma=0.3$. The component centers
\[
\mu_i
=
\sqrt{2}
\left(
\cos\frac{2\pi i}{8},
\sin\frac{2\pi i}{8}
\right),
\quad i=0,\dots,7,
\]
are uniformly distributed on a circle of radius $\sqrt{2}$. To create interleaved high- and low-reward modes, we assign alternating mixture weights $w_i=2$ for even $i$ and $w_i=1$ for odd $i$. The final reward is given by the normalized mixture density, producing a smooth multimodal reward landscape with values in $[0,1]$.

\subsection{Two-Moons}
\begin{figure}[h!]
    \centering
    \includegraphics[width=1.0\linewidth]{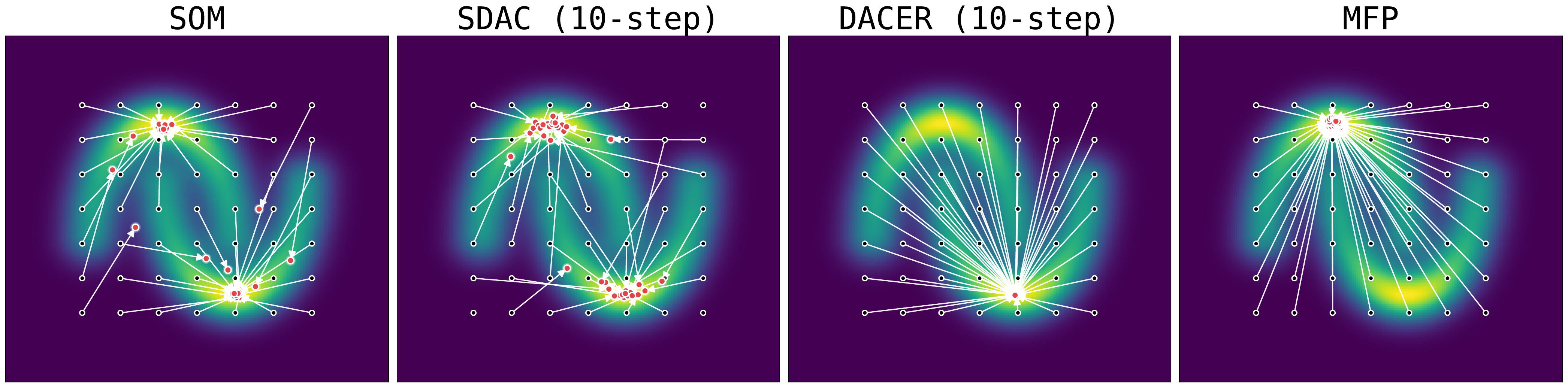}
    \caption{\textbf{Two-Moons Results.}
    Arrow plots from a $7\times 7$ grid of $a_1$ (black) to $a_0$ (red) on the two-moon reward.
    SDAC and DACER with their full 10-step rollouts.
    SOM (Ours) and MFP, both 1-step by design.
    }
    \label{fig:two-moon}
\end{figure}
As shown in \cref{fig:two-moon}, SOM successfully transports actions toward the high-reward modes of the two-moon landscape while preserving coherent and stable transport trajectories. Compared to diffusion-based baselines with iterative denoising steps, SOM achieves comparable mode-seeking behavior using a single-step policy parameterization. In contrast, DACER exhibits more diffuse transport dynamics, whereas MFP occasionally produces less structured trajectories around the multimodal regions.

\subsection{Checkerboard}
\begin{figure}[h!]
    \centering
    \includegraphics[width=1.0\linewidth]{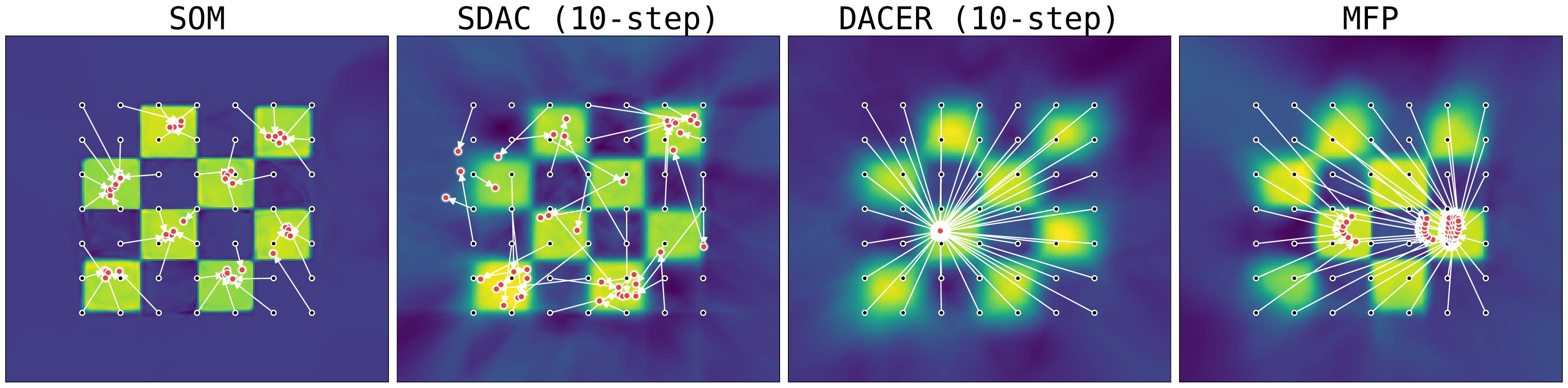}
    \caption{\textbf{Checkerboard results.}
    Arrow plots from a $7\times 7$ grid of $a_1$ (black) to $a_0$ (red) on the checkerboard reward.
    SDAC and DACER with their full 10-step rollouts.
    SOM (Ours) and MFP, both 1-step by design.
    }
    \label{fig:checkerboard}
\end{figure}
As shown in \cref{fig:checkerboard}, SOM accurately transports actions toward high-reward regions in the checkerboard landscape. Compared to iterative diffusion-based baselines, SOM achieves clearer mode separation and more stable convergence behavior under a single-step policy parameterization. DACER exhibits highly diffuse transport patterns, whereas MFP tends to concentrate trajectories toward a limited subset of modes.

\section{Comparison of True and Estimated Scores}
\begin{figure}[h!]
    \centering
    \includegraphics[width=1.0\linewidth, height=3cm, keepaspectratio]{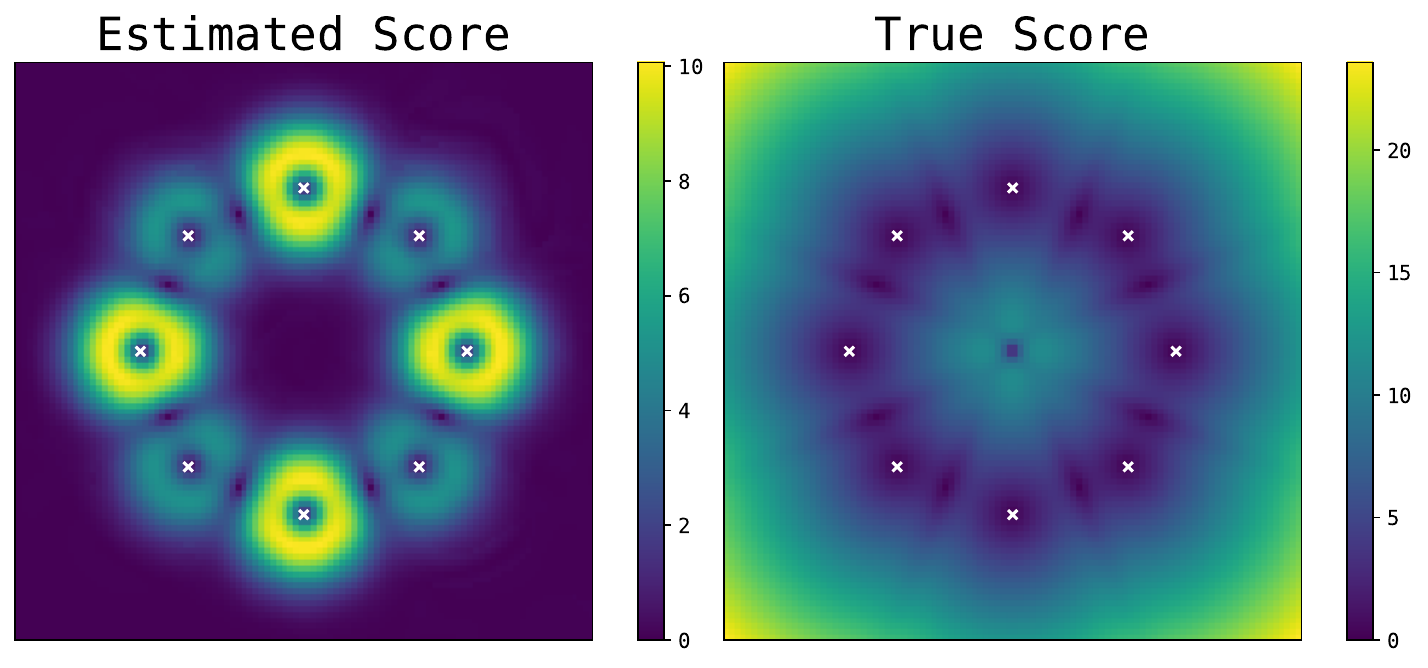}
    \caption{\textbf{Score field comparison on the eight-Gaussian reward.}
    Left: estimated score (max magnitude $\approx 10$).
    Right: ground-truth score (max magnitude $\approx 24$).
    Brighter regions indicate larger gradient magnitude; the eight mode centers are marked with $\times$.
    }
    \label{fig:magnitude_est_vs_true}
\end{figure}
Since the reward landscape is defined as a Gaussian mixture, the corresponding diffusion-perturbed density admits a closed-form expression, making the ground-truth score analytically tractable. We compute this true score and compare it against the score estimated by our method.

The true score function can be derived as follows.
Suppose $p_0(x_0) = \sum_{k=1}^K w_k \mathcal{N}(x_0; \mu_k, \Sigma_k)$ and $p_t(x_t \mid x_0) = \mathcal{N}(x_t; x_0, \sigma_t^2 I)$. 

Marginalizing over $x_0$ gives
\begin{align*}
p_t(x_t)
&=
\int p_t(x_t\mid x_0)
\sum_{k=1}^K
w_k \mathcal N(x_0;\mu_k,\Sigma_k)\,dx_0 \\
&=
\sum_{k=1}^K
w_k
\int
\mathcal N(x_t;x_0,\sigma_t^2 I)
\mathcal N(x_0;\mu_k,\Sigma_k)\,dx_0 \\
&=
\sum_{k=1}^K
w_k
\mathcal N(x_t;\mu_k,\Sigma_k+\sigma_t^2 I),
\end{align*}
where the last equality uses the convolution of two Gaussians.
Applying $\nabla \log f = \nabla f / f$ together with $\nabla_{x_t}\mathcal{N}(x_t;\mu_k,\Sigma_k+\sigma_t^2 I) = \mathcal{N}(x_t;\mu_k,\Sigma_k+\sigma_t^2 I)(\Sigma_k+\sigma_t^2 I)^{-1}(\mu_k-x_t)$ yields
\begin{align*}
\nabla_{x_t}\log p_t(x_t)
&=
\frac{
\sum_{k=1}^K
w_k
\mathcal N(x_t;\mu_k,\Sigma_k+\sigma_t^2 I)
(\Sigma_k+\sigma_t^2 I)^{-1}(\mu_k-x_t)
}{
\sum_{k=1}^K
w_k
\mathcal N(x_t;\mu_k,\Sigma_k+\sigma_t^2 I)
}.
\end{align*}
Letting
\begin{align*}
\rho_k(x_t)
&=
\frac{
w_k
\mathcal N(x_t;\mu_k,\Sigma_k+\sigma_t^2 I)
}{
\sum_{j=1}^K
w_j
\mathcal N(x_t;\mu_j,\Sigma_j+\sigma_t^2 I)
},
\end{align*}
the score reduces to a weighted average of the component-wise scores:
\begin{align*}
\nabla_{x_t}\log p_t(x_t)
&=
\sum_{k=1}^K
\rho_k(x_t)
(\Sigma_k+\sigma_t^2 I)^{-1}
(\mu_k-x_t).
\end{align*}
As shown in \cref{fig:magnitude_est_vs_true}, the ground-truth score (right) is a long-range field: its magnitude grows in the outer region where it pulls samples sharply toward the nearest mode, vanishing only at mode centers, the origin, and inter-mode saddles. 
The estimated score (left), in contrast, is short-range, bright only on a thin annulus around each mode and flat elsewhere, since it fits the reward only locally.
Despite this truncation, the PF-ODE drift in ~\cref{eq:newtargetvel} explains why: the prior-pull term $-\frac{1}{2}\beta(t)\,a_t$ alone contracts outer samples inward at early steps, so an accurate outer-region score is unnecessary; the learned score only needs to be reliable near the modes, where it snaps samples onto the annulus at late steps. 
This division of labor also explains why partial mode collapse does not degrade reward: outer-region accuracy is simply not required.

\section{Robustness under Two Reward Perturbations}
\label{app:robustness_q}
\subsection{Random Gaussian Noise}
At every reward query during training, an independent Gaussian sample with mean $0$ and standard deviation $\sigma$ is added directly to the observed scalar reward, so the same action returns a different value each time it is queried. In this setting $\sigma$ characterizes the magnitude of measurement noise on the reward value itself; it does \emph{not} describe any spatial scale in the action space.

\subsection{Gaussian Bump}
A deterministic Gaussian bump centered at the origin is added to the reward, with peak value equal to $c\cdot Q_{\max}$ (where $c$ is the bias scale and $Q_{\max}$ is the maximum of the original reward). Here $\sigma$ is a \emph{spatial} length scale: it sets the width of the bump in the action space. A larger $\sigma$ produces a wider bump whose influence extends out to the high-reward modes, whereas a smaller $\sigma$ yields a narrow, sharply-peaked bump concentrated near the origin. Crucially, this perturbation is identical on every query (no randomness), and the role of $\sigma$ is fundamentally different from the random Gaussian noise.

\subsection{Radius for Counting Points That Reach A Mode}
The 2D bandit has four high-reward modes lying on the cardinal axes (north, east, south, west) at distance $\sqrt{2}\approx 1.414$ from the origin. For each trajectory endpoint, we count it as having reached a given mode if its Euclidean distance to that mode's center is at most $0.5$. The per-mode counts $\mathrm{N}/\mathrm{E}/\mathrm{S}/\mathrm{W}$ and the total shown below each panel are obtained with this fixed tolerance of $0.5$.

\begin{figure}[h!]
    \centering
    \includegraphics[width=1.0\linewidth]{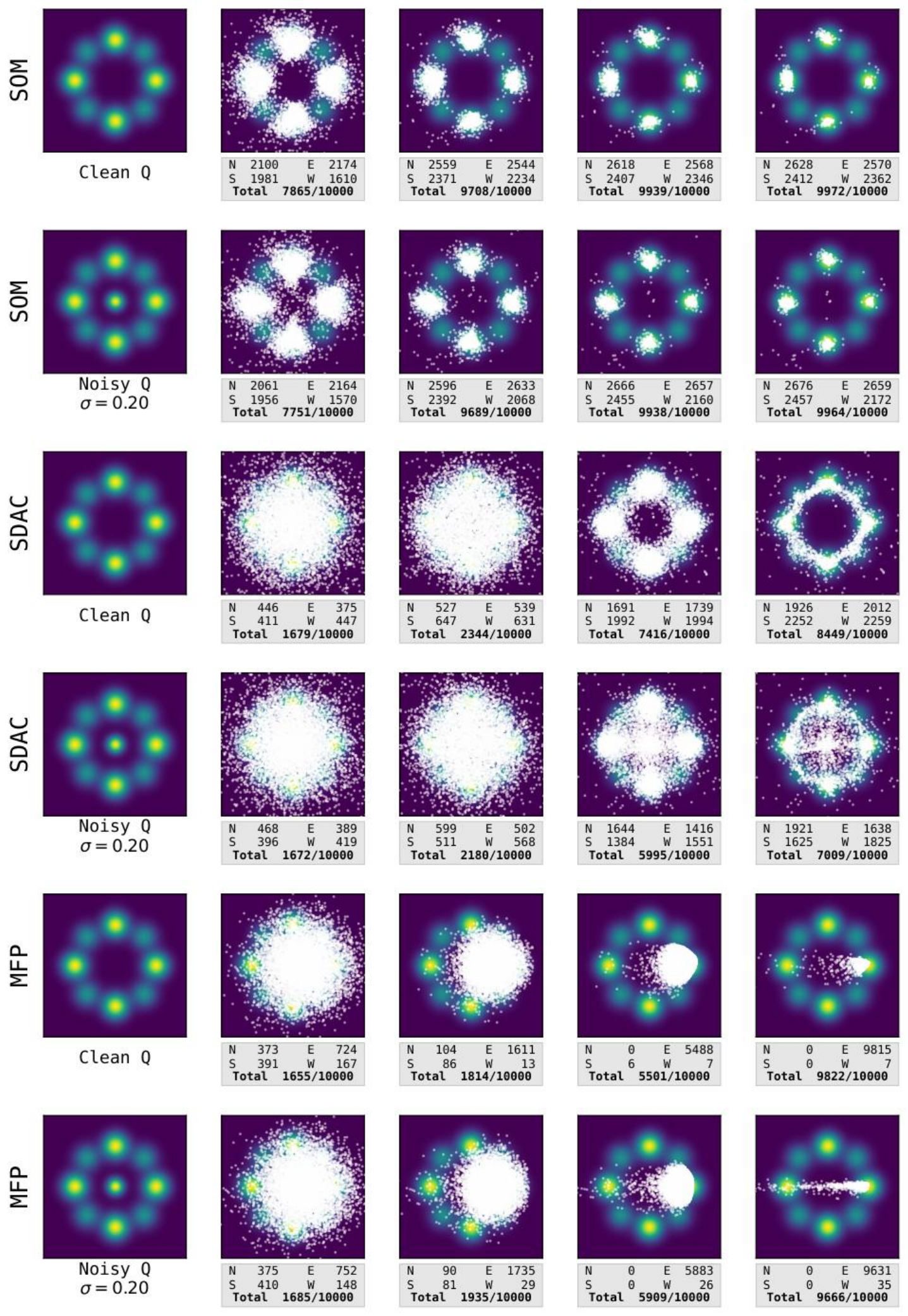}
    \caption{\textbf{Gaussian Bump.}
    }
\end{figure}

\begin{figure}[h!]
    \centering
    \includegraphics[width=1.0\linewidth]{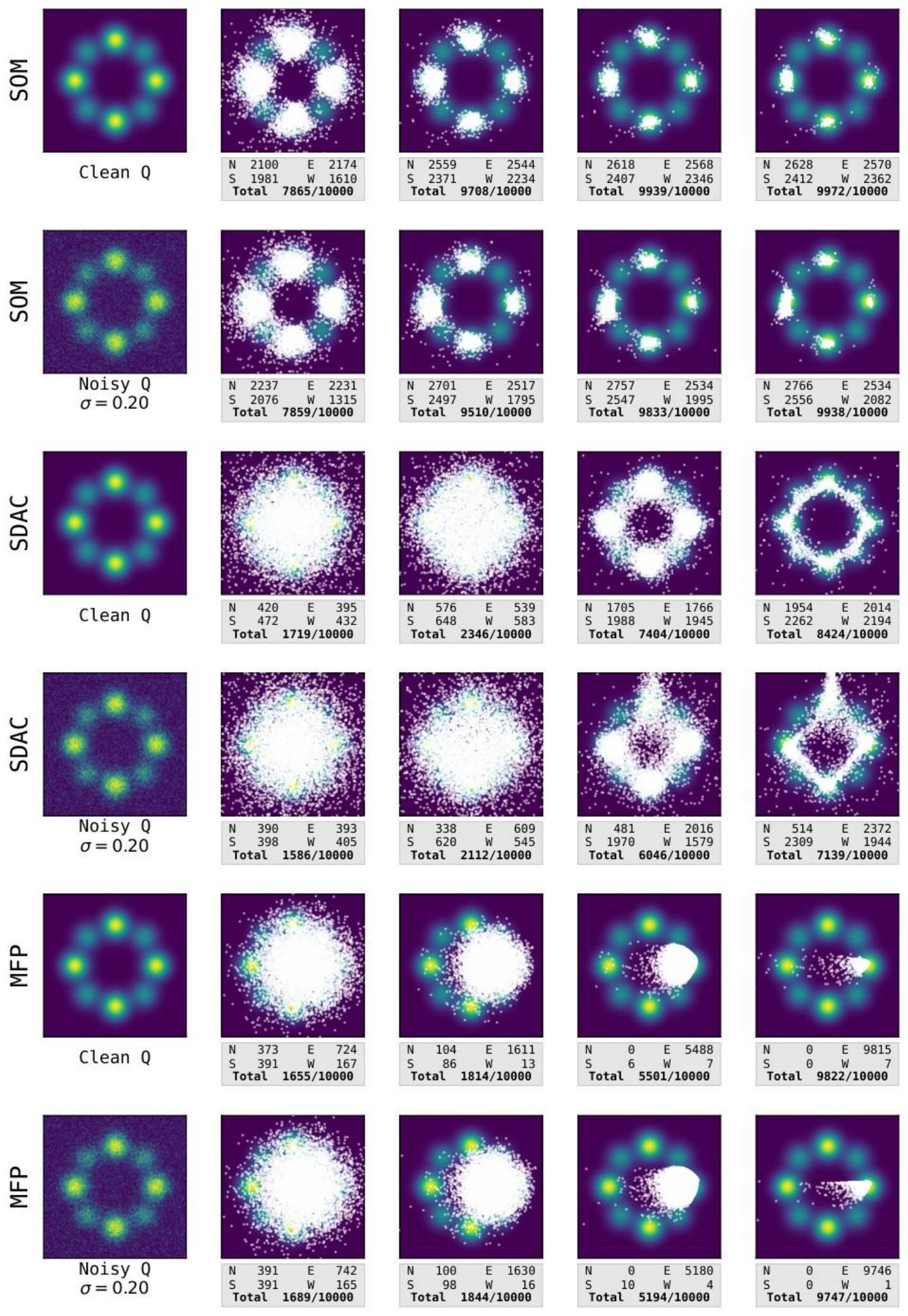}
    \caption{\textbf{Random Gaussian Noise ($\sigma = 2.0$).}
    }
\end{figure}

\begin{figure}[h!]
    \centering
    \includegraphics[width=1.0\linewidth]{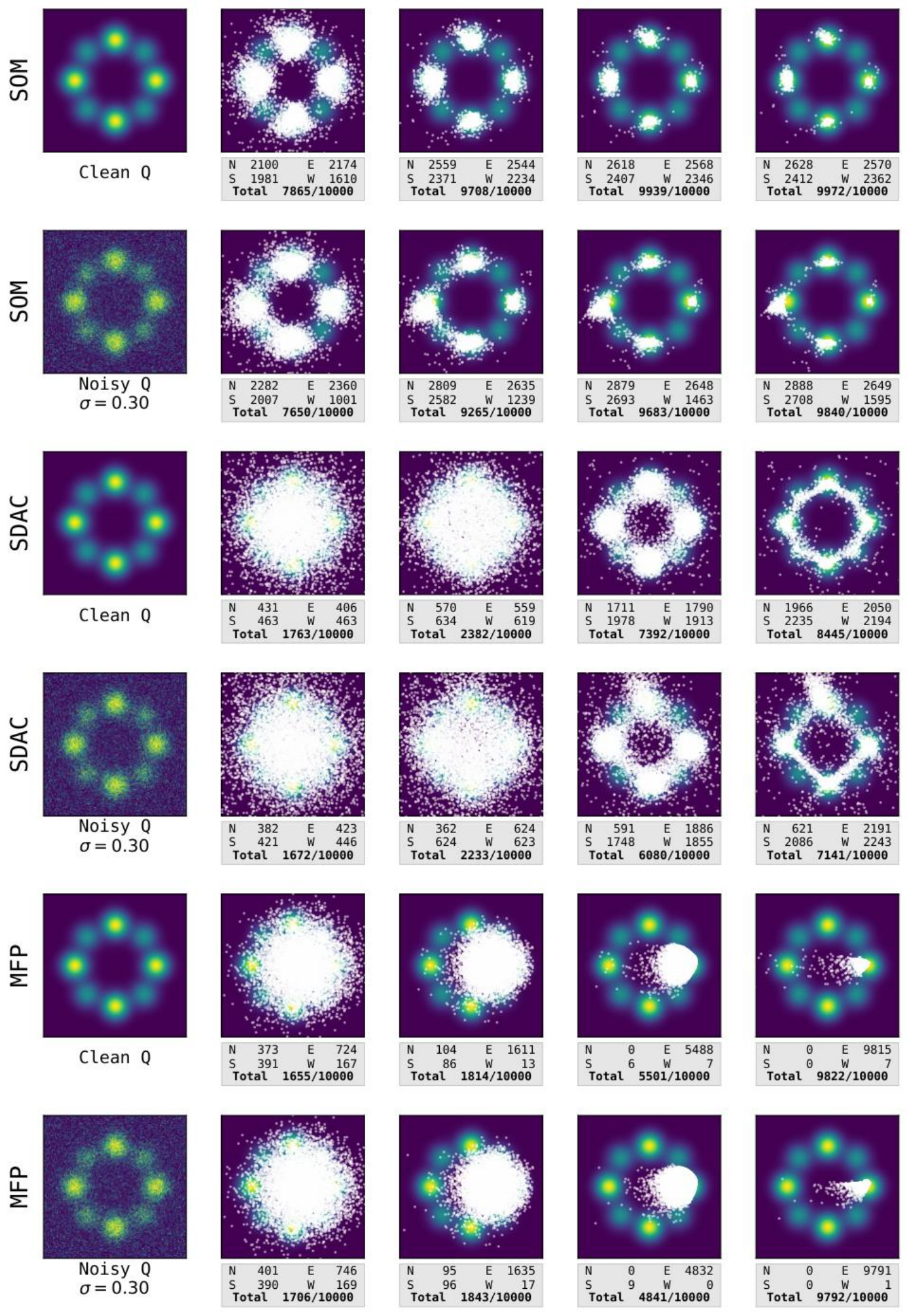}
    \caption{\textbf{Random Gaussian Noise ($\sigma = 3.0$).}
    }
\end{figure}

\section{Implementation Details}
\label{implementation_details}
Note that we conducted all experiments using four NVIDIA RTX 4090 GPUs and AMD Ryzen Threadripper PRO 5995WX CPUs. 

\label{app:implementation}
\begin{table}[h!]
    \centering
    \caption{The overall algorithms' hyperparameter settings.}
    \resizebox{\linewidth}{!}{%
    \begin{tabular}{lcccccccccc}
    \toprule
    \textbf{Hyperparameter}         &\textbf{SOM} & \textbf{SDAC} & \textbf{DPMD}  & \textbf{QSM} & \textbf{DIPO} & \textbf{DACER} & \textbf{QVPO} & \textbf{SAC} &\textbf{PPO} &\textbf{MFP} \\ 
    \midrule
    Replay buffer capacity          & 1e6         & 1e6           & 1e6            & 1e6          & 1e6           & 1e6            & 1e6           & 1e6          & -            & 1e6\\
    Buffer warm-up size             & 1e5         & 3e4           & 3e4            & 3e4          & 3e4           & 3e4            & 3e4           & 3e4          & -            & 3e4 \\
    Batch size                      &  256       & 256           & 256            & 256          & 256           & 256            & 256           & 256          & 4000            & 256 \\
    Discount factor $\gamma$        & 0.99             & 0.99          & 0.99           & 0.99         & 0.99          & 0.99           & 0.99          & 0.99         & 0.995            & 0.99\\
    Target update rate $\tau$       & 0.005            & 0.005         & 0.005          & 0.005        & 0.005         & 0.005          & 0.005         & 0.005        & -            & 0.005\\
    Reward scale                    &  0.2           & 0.2           & 0.2            & 0.2          & 0.2           & 0.2            & 0.2           & 0.2          & 1.0            & 0.2 \\
    Number of hidden layers            &  3           & 3             & 3              & 3            & 3             & 3              & 3             & 3            & 3            & 3 \\
    Number of hidden nodes             &  256           & 256           & 256            & 256          & 256           & 256            & 256           & 256          & 512            & 256 \\
    Activations                     &  GELU           & Mish          & Mish           & ReLU         & Mish          & Mish           & Mish          & GELU         & GELU            & GELU \\
    Diffusion steps                 &  1           & 20            & 20             & 20           & 100           & 20             & 20            & -            & -            & 1\\
    Alpha delay update              &  250           & 250           & 250            & -            & -             & 10,000         & -             & -            & -            & - \\
    Action gradient steps           &  -           & -             & -              & -            & 30            & -              & -             & -            & -            & - \\
    Number of Gaussian distributions   & -            & -             & -              & -            & -             & 3              & -             & -            & -            & - \\
    Number of action samples           &  -           & -             & -              & -            & -             & 200            & -             & -            & -            & - \\
    Noise scale                     &  0.1            & 0.1           & 0.1            & -            & -             & -              & -             & -            & -            & - \\
    Optimizer                       &  Adam           & Adam          & Adam           & Adam         & Adam          & Adam           & Adam          & Adam         & Adam            & Adam \\
    Actor learning rate             &  1e-4           & 3e-4          & 3e-4           & 3e-4         & 3e-4          & 3e-4           & 3e-4          & 3e-4         & 1e-4            & 1e-4\\
    Critic learning rate            &  1e-4           & 3e-4          & 3e-4           & 3e-4         & 3e-4          & 3e-4           & 3e-4          & 3e-4         & 1e-3            & 1e-4\\
    Alpha learning rate             &  7e-3           & 7e-3          & 7e-3           & -            & -             & 3e-2           & -             & 3e-4         & -            & -\\
    Target entropy                  &  -0.9 $\cdot$ dim($\mathcal{A}$)           & -0.9 $\cdot$ dim($\mathcal{A}$) & -0.9 $\cdot$ dim($\mathcal{A}$) & - & - & -0.9 $\cdot$ dim($\mathcal{A}$) & - & -dim($\mathcal{A}$) & - \\
    Number of particles for boN    &   32          & 32            & 32             & 32           & 1             & 1              & 32            & 1            &  -           & 32 \\ 
    \bottomrule
    \end{tabular}%
    }
    \label{tab:hyperparams_baselines}
\end{table}

\begin{table}[h!]
\centering
\caption{The PPO algorithm's hyperparameter settings.}
\vspace{4pt}
\begin{tabular}{lc}
\toprule
\textbf{Hyperparameter} & \textbf{Setting} \\
\midrule
GAE $\lambda$ & 0.95 \\
PPO clip $\epsilon$ & 0.2 \\
Target KL & 0.01 \\
Policy update steps & 80 \\
Value update steps & 80 \\
\bottomrule
\end{tabular}
\end{table}

\end{document}